%% file: parallel-astar.tex
\newif\ifhandbook
\handbookfalse
\ifhandbook
\documentclass[graybox]{svmult}

\usepackage{mathptmx}       
\usepackage{helvet}         
\usepackage{courier}        
\usepackage{type1cm}        
\usepackage{makeidx}         
\usepackage{graphicx}        
\usepackage{multicol}        
\usepackage[bottom]{footmisc}

\makeindex             

\else 
\documentclass{article}
\usepackage{authblk}
\usepackage[margin=1in]{geometry}
\usepackage{amsthm}
\newtheorem{definition}{Definition}
\newtheorem{theorem}{Theorem}

\fi

\usepackage{amssymb}
\usepackage{amsmath}
\usepackage{comment}
\usepackage{tabularx}
\usepackage{subfig}
\usepackage{relsize}
\usepackage{tikz}

\usepackage{hyperref}

\newcommand{\fromaij}[1]{} 
\newcommand{\fromjfjournal}[1]{} 

 \newcommand{\todoadi}[1]{}
 \newcommand{\todoyuu}[1]{}

 \newcommand{\todo}[1]{}

\newcommand{\sharedopen}[0]{$OPEN_{shared}\;$}
\newcommand{\sharedclosed}[0]{$CLOSED_{shared}\;$}

\newcommand{\citeyear}[1]{\cite{#1}}
\newcommand{\citeauthor}[1]{\cite{#1}}

\newcommand{\term}[1]{$\mathit{#1}$} 
\newcommand{\hda}{H\!D\!A^*\!}

\newcommand{\GRAZHDAS}{\term{\hda[Z, A_{feature}/DTG_{sparsity}]}}

\newcommand{\FAZHDA}{\term{\hda[Z, A_{feature}/DTG_{fluency}]}} 
\newcommand{\GAZHDA}{\term{\hda[Z, A_{feature}/DTG_{greedy}]}}

\newcommand{\AZHDA}{\term{\hda[Z, A_{feature}]}} 

\newcommand{\DAHDA}{\term{\hda[Z, A_{state}/SDD_{dynamic}]}}
\newcommand{\OZHDA}{\term{\hda[Z_{operator}]}} 

\newcommand{\AHDAP}{\term{\hda[P, A_{state}]}}
\newcommand{\AHDAPSSD}{\term{\hda[P, A_{state}/SDD]}}

\newcommand{\AHDAZ}{\term{\hda[Z, A_{state}]}}
\newcommand{\AHDAZSSD}{\term{\hda[Z, A_{state}/SDD]}}

\newcommand{\HWD}{\term{\hda[Hyperplane]}}
\newcommand{\PHDA}{\term{\hda[P]}}
\newcommand{\ZHDA}{\term{\hda[Z]}}

\newcommand{\apDAHDA}{A}

\usepackage[ruled,boxed,linesnumbered,noend]{algorithm2e}
\SetKw{KwDownTo}{down to}

\begin{document}

\ifhandbook
\title*{Parallel A* For State-Space Search}
\author{Alex Fukunaga, Adi Botea, Yuu Jinnai, and Akihiro Kishimoto}
\institute{
Alex Fukunaga \at The University of Tokyo, Tokyo, Japan, \email{fukunaga@idea.c.u-tokyo.ac.jp}
\and 
Adi Botea \at IBM Research, Dublin, Ireland,  \email{ADIBOTEA@ie.ibm.com}
\and 
Yuu Jinnai \at The University of Tokyo, Tokyo, Japan, \email{ddyuudd@gmail.com}
\and
Akihiro Kishimoto \at IBM Research, Dublin, Ireland,  \email{AKIHIROK@ie.ibm.com}
}


\else   

\title{A Survey of Parallel A*}
\author[*]{Alex Fukunaga}
\author[**]{Adi Botea}
\author[*]{Yuu Jinnai}
\author[**]{Akihiro Kishimoto}
\affil[*]{Graduate School of Arts and Sciences, The University of Tokyo}
\affil[**]{IBM Research, Ireland}
\fi

\maketitle

\abstract{
  A* is a best-first search algorithm for finding optimal-cost paths
  in graphs.  A* benefits significantly from parallelism because in
  many applications, A* is limited by memory usage, so distributed
  memory implementations of A* that use all of the aggregate memory on
  the cluster enable problems that can not be solved by
  serial, single-machine implementations to be solved.  We survey approaches to
  parallel A*, focusing on decentralized approaches to A* which
  partition the state space among processors. We also survey approaches to parallel, limited-memory variants of A* such as parallel IDA*.
}


\input intro

\input preliminaries
\input parallel-bfs-algs

\input hash-based-dec-astar

\input dec-search-str-part

\input hash-functions-hdastar
\input portofolios-astar
\input mem-lim-astar

\input astar-cloud

\input astar-gpus
\input other-approaches

\ifhandbook
\begin{acknowledgement}
\else
\section*{Acknowledgements}
\fi
This work was supported in part by JSPS KAKENHI grants 25330253 and 17K00296.

\ifhandbook
\end{acknowledgement}
\fi

\bibliographystyle{spmpsci}
\bibliography{ref-jf16}

\ifhandbook
\printindex
\fi

\end{document}

%% file: intro.tex
\section{Introduction}


This chapter surveys parallel A* for state-space search.
State-space search \index{state-space search} is a very general approach to solving a broad class of problems,
such as robot planning problems,
domain-independent AI planning,
solving puzzles,
and multiple sequence alignment problems in computational biology.

Solving a problem with state-space search involves defining the
\emph{state space} as a graph where nodes represent states and
edges represent actions (transitions) between states.
The task is to find 
a sequence of actions which
transforms a given initial state into a state that satisfies some goal conditions.
In other words, finding a solution boils down to finding
a path from the initial state to a goal state.
The quality of a solution is typically measured in terms of the 
total cost of the path. The smaller the cost, the better the solution.
Optimal search aims at finding a minimal-cost solution.
We formally define concepts such as state spaces, state-space search problems,
and (optimal) solutions after the next example.

\begin{figure}
\centering
\begin{tabular}{cc}

\begin{tikzpicture}
\fill [white] (0.0,7.2) rectangle (0.8,8.0);
\draw [black] (0.0,7.2) rectangle (0.8,8.0);
\fill [white] (0.8,7.2) rectangle (1.6,8.0);
\draw [black] (0.8,7.2) rectangle (1.6,8.0);
\fill [white] (1.6,7.2) rectangle (2.4,8.0);
\draw [black] (1.6,7.2) rectangle (2.4,8.0);
\fill [white] (2.4,7.2) rectangle (3.2,8.0);
\draw [black] (2.4,7.2) rectangle (3.2,8.0);
\fill [white] (0.0,6.4) rectangle (0.8,7.2);
\draw [black] (0.0,6.4) rectangle (0.8,7.2);
\fill [white] (0.8,6.4) rectangle (1.6,7.2);
\draw [black] (0.8,6.4) rectangle (1.6,7.2);
\fill [white] (1.6,6.4) rectangle (2.4,7.2);
\draw [black] (1.6,6.4) rectangle (2.4,7.2);
\fill [white] (2.4,6.4) rectangle (3.2,7.2);
\draw [black] (2.4,6.4) rectangle (3.2,7.2);
\fill [white] (0.0,5.6) rectangle (0.8,6.4);
\draw [black] (0.0,5.6) rectangle (0.8,6.4);
\fill [white] (0.8,5.6) rectangle (1.6,6.4);
\draw [black] (0.8,5.6) rectangle (1.6,6.4);
\fill [white] (1.6,5.6) rectangle (2.4,6.4);
\draw [black] (1.6,5.6) rectangle (2.4,6.4);
\fill [white] (2.4,5.6) rectangle (3.2,6.4);
\draw [black] (2.4,5.6) rectangle (3.2,6.4);
\fill [white] (0.0,4.8) rectangle (0.8,5.6);
\draw [black] (0.0,4.8) rectangle (0.8,5.6);
\fill [white] (0.8,4.8) rectangle (1.6,5.6);
\draw [black] (0.8,4.8) rectangle (1.6,5.6);
\fill [white] (1.6,4.8) rectangle (2.4,5.6);
\draw [black] (1.6,4.8) rectangle (2.4,5.6);
\fill [white] (2.4,4.8) rectangle (3.2,5.6);
\draw [black] (2.4,4.8) rectangle (3.2,5.6);
\fill [white] (3.2,4.8) rectangle (4.0,5.6);
\draw [black] (2.4,4.8) rectangle (3.2,5.6);
\draw (0.4,7.6) node{\LARGE 15};
\draw (1.2,7.6) node{\LARGE 2};
\draw (2.0,7.6) node{\LARGE 8};
\draw (2.8,7.6) node{\LARGE 7};

\draw (0.4,6.8) node{\LARGE 1};
\draw (1.2,6.8) node{\LARGE 6};
\draw (2.0,6.8) node{\LARGE 9};
\draw (2.8,6.8) node{\LARGE 11};

\draw (0.4,6.0) node{\LARGE 13};
\draw (1.2,6.0) node{\LARGE 12};
\draw (2.8,6.0) node{\LARGE 4};

\draw (0.4,5.2) node{\LARGE 10};
\draw (1.2,5.2) node{\LARGE 5};
\draw (2.0,5.2) node{\LARGE 3};
\draw (2.8,5.2) node{\LARGE 14};
\end{tikzpicture}

&

\begin{tikzpicture}
\fill [white] (0.0,7.2) rectangle (0.8,8.0);
\draw [black] (0.0,7.2) rectangle (0.8,8.0);
\fill [white] (0.8,7.2) rectangle (1.6,8.0);
\draw [black] (0.8,7.2) rectangle (1.6,8.0);
\fill [white] (1.6,7.2) rectangle (2.4,8.0);
\draw [black] (1.6,7.2) rectangle (2.4,8.0);
\fill [white] (2.4,7.2) rectangle (3.2,8.0);
\draw [black] (2.4,7.2) rectangle (3.2,8.0);
\fill [white] (0.0,6.4) rectangle (0.8,7.2);
\draw [black] (0.0,6.4) rectangle (0.8,7.2);
\fill [white] (0.8,6.4) rectangle (1.6,7.2);
\draw [black] (0.8,6.4) rectangle (1.6,7.2);
\fill [white] (1.6,6.4) rectangle (2.4,7.2);
\draw [black] (1.6,6.4) rectangle (2.4,7.2);
\fill [white] (2.4,6.4) rectangle (3.2,7.2);
\draw [black] (2.4,6.4) rectangle (3.2,7.2);
\fill [white] (0.0,5.6) rectangle (0.8,6.4);
\draw [black] (0.0,5.6) rectangle (0.8,6.4);
\fill [white] (0.8,5.6) rectangle (1.6,6.4);
\draw [black] (0.8,5.6) rectangle (1.6,6.4);
\fill [white] (1.6,5.6) rectangle (2.4,6.4);
\draw [black] (1.6,5.6) rectangle (2.4,6.4);
\fill [white] (2.4,5.6) rectangle (3.2,6.4);
\draw [black] (2.4,5.6) rectangle (3.2,6.4);
\fill [white] (0.0,4.8) rectangle (0.8,5.6);
\draw [black] (0.0,4.8) rectangle (0.8,5.6);
\fill [white] (0.8,4.8) rectangle (1.6,5.6);
\draw [black] (0.8,4.8) rectangle (1.6,5.6);
\fill [white] (1.6,4.8) rectangle (2.4,5.6);
\draw [black] (1.6,4.8) rectangle (2.4,5.6);
\fill [white] (2.4,4.8) rectangle (3.2,5.6);
\draw [black] (2.4,4.8) rectangle (3.2,5.6);
\draw (0.4,7.6) node{\LARGE 1};
\draw (1.2,7.6) node{\LARGE 2};
\draw (2.0,7.6) node{\LARGE 3};
\draw (2.8,7.6) node{\LARGE 4};

\draw (0.4,6.8) node{\LARGE 5};
\draw (1.2,6.8) node{\LARGE 6};
\draw (2.0,6.8) node{\LARGE 7};
\draw (2.8,6.8) node{\LARGE 8};

\draw (0.4,6.0) node{\LARGE 9};
\draw (1.2,6.0) node{\LARGE 10};
\draw (2.0,6.0) node{\LARGE 11};
\draw (2.8,6.0) node{\LARGE 12};

\draw (0.4,5.2) node{\LARGE 13};
\draw (1.2,5.2) node{\LARGE 14};
\draw (2.0,5.2) node{\LARGE 15};
\end{tikzpicture}

\end{tabular} 

\caption{An instance of the sliding-tile puzzle with 15 tiles.
Left: initial state. Right: goal state.
\label{eleven:fig:15-puzzle}
}
\end{figure}

Consider the simple sliding-tile problem, where an $n \times n$ board is occupied by $n^2-1$ tiles and a ``blank'' space.
As there are $n^2-1$ tiles, the problem is also called the $n^2-1$ puzzle.
Figure \ref{eleven:fig:15-puzzle} illustrates a 15 puzzle instance.
Given an initial state where the tiles are out of order, the task is to find a sequence of actions which results in the goal state where the tiles are in order.
The actual, physical problem is played by moving one of the tiles currently adjacent to the blank space into the blank space.
As this is equivalent to moving the blank space, it is customary to treat the problem as having four actions ({\it up, down, left}, and {\it right}), which move the blank space.
The most common variant of this problem in the AI literature requires finding the solution with the minimal number of moves.
The problem of finding optimal solution for the sliding-tile puzzle has been used as a standard benchmark problem in the AI search algorithm literature because of the simplicity of the problem description and the difficulty of finding an optimal solution.
Although puzzles of small sizes, such as $3 \times 3$ and $4 \times 4$,
can be solved fairly easily, 
a $5 \times 5$ version of this puzzle has $(5 \times 5)!/2 \approx 7.76 \times 10^{24}$ possible configurations, providing a challenging benchmark for search algorithms.

This survey is structured as follows.
First, in Section \ref{eleven:sec:astar}, we give a formal definition of state-space search, and review the A* algorithm \cite{hart68formal},
which is the standard, baseline  approach for optimally solving state-space search problems.
Next, in Section \ref{eleven:sec:parallel-best-first-search}, we give an overview of the parallel-search related overheads which pose the fundamental challenges in parallelizing A*, and review the two basic approaches to parallelizing A*: the centralized and decentralized approaches.
Then, in Section \ref{eleven:sec:hash-based-decentralized}, we describe hash-based work distribution, the class of algorithms which is the current, state-of-the-art approach for parallelizing A* both on single, shared-memory multi-core machines as well as large-scale clusters. Hash-based work distribution handles both load balancing and efficient detection of duplicate states. 
Section \ref{eleven:sec:structure-based} reviews structure-based search space partitioning, which is another approach to decentralized search based on the concept of a duplicate detection scope which allows minimization of communications among processors.
Section \ref{eleven:sec:hash-functions} surveys the various approaches to implementing hashing strategies for hash-based work distribution, including recent approaches which integrate key ideas from structure-based search space partitioning.
Parallel portfolios (meta-solvers that combine multiple problem solvers) which include A*-based solvers as a component are reviewed in Section \ref{eleven:sec:portfolio-astar}.
A fundamental limitation of A* is that it can exhaust memory on hard problems, resulting in failure to solve the problem.
Limited-memory variants of A* such as IDA* \cite{korf:85a} overcome this limitation (at the cost of some search efficiency). We survey parallel, limited-memory A* variants in Section \ref{eleven:sec:memory-limited}. 
With the emergence of cloud environments offering virtually unlimited available resources (at a cost), another approach to addressing the A* memory usage problem is to simply use more machines. Section \ref{eleven:sec:cloud} describes resource allocation strategies for parallel A* that are efficient with respect to cost and runtime. Recently, graphics processing units (GPUs) with thousands of cores have become widely used. Since GPUs have afundamentally different architecture compared to traditional CPUs, this  provides new challenges for parallelizing A*. Section \ref{eleven:sec:gpu} describes recent work on parallel A* variants for GPUs.
Finally, Section \ref{eleven:sec:misc-approaches} reviews other approaches to parallel state-space search.

%% file: preliminaries.tex
\section{Preliminaries: Review of A*}
\label{eleven:sec:astar}

This section provides preliminary and background material for the rest of this 
\ifhandbook
chapter.
\else
paper.
\fi
We first formally define state-space search, and then present the A* search algorithm.

The formal definitions presented below 
are adapted from
Edelkamp and Schroedl's 
textbook on heuristic search \cite{Edelkamp:2010:HST:1875144}.
%
\begin{definition}[State space problem]
A {\emph state space problem} $P=(S,A,s_0,T)$ is defined by a set of states $S$, an initial state $s_0 \in S$, a set of goal states $T \subset S$, and a finite set of actions $A={a_1,...,a_m}$ where each $a_i:S \rightarrow S$ transforms a state into another state.
\end{definition}

For the sliding-tile puzzle, the state space problem formulation consists of the states $S$, where each state corresponds to a unique configuration of the tiles, $s_0$ is the given initial configuration, $T$ is a singleton set whose sole member is the configuration with the tiles in the correct order, and $A$ corresponds to the transitions between tile configurations.

\begin{definition} 
	[State Space Problem Graph] 
	A problem graph $G=(V,E,s_0,T)$ for the state space problem $P=(S,A,s_0,T)$ is defined by $V=S$ as the set of nodes, $s_0 \in S$ as the initial node, $T$ as the set of goal nodes, and $E \subset V \times V$ as the set of edges that connect nodes to nodes with $(u,v) \in E$ if and only if there exists an $a \in A$ with $a(u) = v$.
\end{definition}

\begin{definition} 
	[Solution] A solution $\pi = (a_1,...,a_k)$ is an ordered sequence of actions $a_i \in A, i \in {1,...,k}$ that transforms the initial state $s_0$ into one of the goal states $t \in T$; that is, there exists a sequence of states $u_i \in S, i \in {0,...,k}$, with $u_0  = s_0, u_k=t$, and $u_i$ is the outcome of applying $a_i$ to  $u_{i-1}$, $i \in {1,...,k}$.
\end{definition}

In some problems, such as the sliding-tile puzzle,
all actions have the same cost.
In other problems, however, different actions can have different costs.
Take pathfinding on a gridmap for example.
Pathfinding refers to computing a path for a mobile
agent, such as a robot or a character in a game,
from an initial location to a target location.
Gridmaps are a popular approach to discretizing the environment
of the agent (e.g., a game map) into a search graph.
A gridmap is a two-dimensional array where a cell is either traversable
or blocked by an obstacle.
The mobile agent occupies exactly one traversable cell at a time.
In creating the problem graph,
all traversable cells become states in the state space (equivalently, 
nodes in the problem graph).
Two adjacent traversable states are connected with an edge.
On a so-called 8-connected gridmap, or octile gridmap,
adjacency relations are defined in 8 directions,
four straight and four diagonal.
Straight edges have a cost of 1, and diagonal edges
have a cost of $\sqrt{2}$.
State spaces where actions have different costs are
called weighted state spaces.

\begin{definition} 
	A weighted state space problem $P=(S,A,s_0,T,w)$, where $w$ is a cost function $w: A \rightarrow \mathbb{R}$. The cost of a path consisting of actions $a_1,...,a_n$ is defined as $\sum^n_{i=1} w(a_i)$. In
	For a weighted state space problem, there is a corresponding {\em weighted problem graph} $G=(V,E,s_0,T,w)$, 
	where $w$ is extended to $E \rightarrow \mathbb{R}$ in the straightforward way. 
	The graph is uniformly weighted if $w(u,v)$ is constant for all $(u,v) \in E$. 
	The weight or cost of a path $\pi = (v_0,...,v_k)$ is defined as $w(\pi) = \sum^k_{i=1} w(v_{i-1},v_i)$.
\end{definition}

\begin{definition} A solution from $s_0$ to a given goal state $v$ is \emph{optimal} if its weight is minimal among all paths between $s_0$ and $v$.
\end{definition}

A state space is \emph{undirected} if for every action from a state $u$ to a state $v$,
there exists an action from state $v$ to state $u$.
Otherwise, the state space is \emph{directed}.
For example, the sliding tile puzzle has an undirected state space,
as every action can be reversed.
On the other hand, planning an itinerary on a road map with 
one-way roads is a directed state space problem.

In some domains, the problem graph is sufficiently small to 
fit into the memory of the computer.
In such cases, the search graph can be defined \emph{explicitly},
enumerating all nodes and edges.
Pathfinding on gridmaps is a typical example of a problem where
the search graph can be defined explicitly.
In many other problems, the search graph is very large,
much larger than can fit into the memory of a modern computer.
Examples include puzzles such as the Rubik's cube and the 
sliding tile puzzle, as well as many benchmark domains
in AI domain-independent planning.
In such cases, the search graph is defined \emph{implicitly}.
Defining a search graph implicitly requires
three key ingredients:
a specification of the initial state,
a method for recognizing goal nodes,
and a method for expanding any node $v \in V$.
Expanding a node $v$ refers to generating all nodes
$u$ such that $(v,u)$ is an edge in the problem graph.

\begin{definition} 
	In an {\em implicit state space graph}, we have an initial node $s_0 \in V$, a set of goal nodes determined by a predicate $Goal: V \rightarrow \mathbb{B} = \{\mbox{false},\mbox{true}\}$, and a node expansion procedure $Expand: V \rightarrow 2^V$.
\end{definition}

Defining a graph implicitly allows us to generate portions of the search
graph on demand, as a given search algorithm needs to explore
new parts of the search graph.

\subsection{The A* Algorithm}
\index{A* search algorithm}
\begin{algorithm}
	
	Initialize $\mbox{OPEN}$ to $\{s_0\}$\;
    \While {$\mbox{OPEN} \neq \emptyset$} {
	    Get and remove from OPEN a node $n$ with a smallest $f(n)$\;
	    Add $n$ to CLOSED\;
	    \If {$n$ is a goal node} {
	    	Return solution path from $s_0$ to $n$\;
	    }
	    \For {every successor $n'$ of $n$} {
	    	$g_1 = g(n) + c(n,n')$\;
	    	\If {$n' \in \mbox{CLOSED}$} {
	    		\If {$g_1 < g(n')$} {
	    			Remove $n'$ from CLOSED and add it to OPEN\;
	    		} \Else {
	    		    Continue\;
		    	}
	    	} \Else {
		    	\If {$n' \notin \mbox{OPEN}$} {
		    		Add $n'$ to OPEN\;
		    	}
		    	\ElseIf {$g_1 \geq g(n')$} {
		    		Continue\;
		    	}
		    }
   			Set $g(n') = g_1$\;
   			Set $f(n') = g(n') + h(n')$\;
   			Set $\mbox{parent}(n') = n$\;
	    }
    }
    Return failure (no path exists)\;
	
	\caption{A*}
	\label{eleven:fig:astar-pseudocode}
\end{algorithm}

Most of the  parallel state-space search algorithms presented in this chapter are based on the serial algorithm A*~\cite{hart68formal}.
A* is a best-first search algorithm whose pseudocode is illustrated in Algorithm~\ref{eleven:fig:astar-pseudocode}.
A* keeps two sets of nodes, called the OPEN list and the CLOSED list.
The CLOSED list is the set of expanded nodes. Recall that expanding a node refers
to generating its successors. The OPEN list contains the nodes
that have been generated and are waiting to be expanded. 
At each iteration of the main while loop shown in the pseudocode, A* selects for expansion a node from the OPEN list,
with the smallest $f$-value. The $f$-value of a node $n$ is defined as $f(n) = g(n) + h(n)$.
The $g(n)$ value is the cost of the best known path from the root node $s_0$ to the current node $n$.
The $h(n)$ value, called the heuristic evaluation of $n$, is an estimation of the cost from $n$ to a closest goal node.
As such, $f(n)$ estimates the cost of a shortest solution passing through $n$.

A heuristic function $h$ \index{heuristic function} is {\em admissible} if $h(n) \leq C^*(n)$, where $C^*(n)$
is the cost of the minimal path from $n$ to some goal, i.e., $h$ is a lower bound on $C^*$.
A heuristic function is \emph{consistent} (or \emph{monotonic}) if $h(n) \leq c(n,n') + h(n')$ for all nodes
$n$ and $n'$ such that $n'$ is a successor of $n$; and $h(t) = 0$ for all goal nodes $t$.
A consistent heuristic is also admissible.
With a consistent heuristic is used, nodes are never reopened (i.e., moved from 
CLOSED to OPEN as shown at line 11 of Algorithm~\ref{eleven:fig:astar-pseudocode}).
In other words, every node is expanded at most once.
This allows us to simplify the algorithm when consistent heuristics are used,
replacing lines 10--13 with a ``Continue'' statement.

An algorithm is {\em complete} if it terminates and returns a solution whenever a solution exists.
An algorithm is \emph{admissible} \index{admissible heuristic} if it always returns an optimal solution whenever a solution exists.

\begin{theorem}
	A* is complete on both finite and infinite graphs \cite{Pearl84}. 
\end{theorem}

\begin{theorem}
	If $h$ is an admissible function, then A* using $h$ is admissible \cite{hart68formal}. 
\end{theorem}

Besides producing optimal solutions, another powerful feature of A* is that it is
an \emph{efficient} algorithm in terms of the number of node expansions performed.
For simplicity, assume that a consistent heuristic is used, to ensure that
there are no re-expansions. A* expands all nodes $n$ with $f(n) < C^*$, where $C^*$ is the optimal solution cost.
It also expands some of the nodes $n$ with $f(n) = C^*$, and no node with $f(n) > C^*$.
A* is efficient because any other admissible algorithm using the same 
knowledge (e.g., the same heuristic $h$) must expand 
all nodes $n$ with $f(n) < C^*$. The reason is that, according to the knowledge 
available, a node $n$ with $f(n) < C^*$ might belong to a solution with a smaller cost than $C^*$.
Unless extra information is available (e.g., pruning based on symmetries in the state space)
the node $n$ has to be expanded to explore whether a better solution can be found.

Besides being a powerful property of serial A*,
the efficiency of A* in terms of node expansions has a special significance to parallel best-first search.
It allows us to evaluate the efficiency of a parallel search algorithm,
such as HDA*~\cite{kishimotofb09}, even in very difficult instances where serial A* fails and therefore a direct comparison
of the node expansions between HDA* and A* is not possible.
The idea is to measure the number of expanded nodes $n$ with $f(n) < C^*$ as a fraction of all expanded nodes.
If the fraction is close to 1, then the instance at hand is solved quite efficiently~\cite{kishimotofb13}.

%% file: parallel-bfs-algs.tex
\section{Parallel Best-First Search Algorithms}
\label{eleven:sec:parallel-best-first-search}

Parallelization of A* heuristic search is important due to two reasons.
First, effective parallelization is necessary in order to obtain good speedup on multi-core processors. 
However, in the case of parallelization on a cluster consisting of many machines, parallelization offers another benefit which is at least as important as speedup, which is increased aggregate memory.
A* memory usage continuously increases during the run, as it must keep all expanded nodes in memory in order to guarantee the soundness (optimality of solution) and completeness of the algorithm.
Running parallel A* on  a cluster of machines makes the entire aggregate memory of the cluster available to A*. This allows parallel A* to solve problem instances that would not be solvable at all  on a single machine (using the same heuristic function).
This offers a fundamental benefit to parallelization of A*, and perhaps makes parallelization of A* an even more pressing concern than for other search algorithms.

In this section, we first describe the major technical challenges that must be addressed in parallel A*, and then describe the two basic approaches to parallelization of A*: centralized and decentralized parallelization.

\subsection{Parallel Overheads}
\label{eleven:sec:parallel-overheads}

\fromaij{background}
Efficient implementation of parallel 
search algorithms is challenging 
due to several types of overhead.
{\it Search overhead (SO)} occurs when 
a parallel implementation of a search algorithm expands (or generates)
more states than a serial implementation.
The main cause of search overhead is partitioning of the search space
among processors,
which has the side effect that access to non-local information is restricted.
For example, sequential A* can terminate immediately after
a solution is found, because it is guaranteed to be optimal. 
In contrast, when a parallel A* algorithm finds a (first)
solution at some processor, it is not necessarily a globally  optimal solution.
A better solution which uses nodes being processed in some other processor might exist.

{\it Synchronization overhead} is the idle time wasted 
when some processors have to wait for the others to
reach synchronization points.
For example, in a shared-memory environment, the idle time can be
caused by mutual exclusion locks on shared data.
Finally, {\it communication overhead (CO)} refers to the cost of 
inter-process information exchange.  In a distributed-memory environment, this includes the cost of sending a message from one processor to another over a network. 
Even in a shared-memory environment, there are overheads associated with moving work from one work queue to another.



\begin{figure}[htb]
	\centering
	\includegraphics[width=1.0\linewidth]{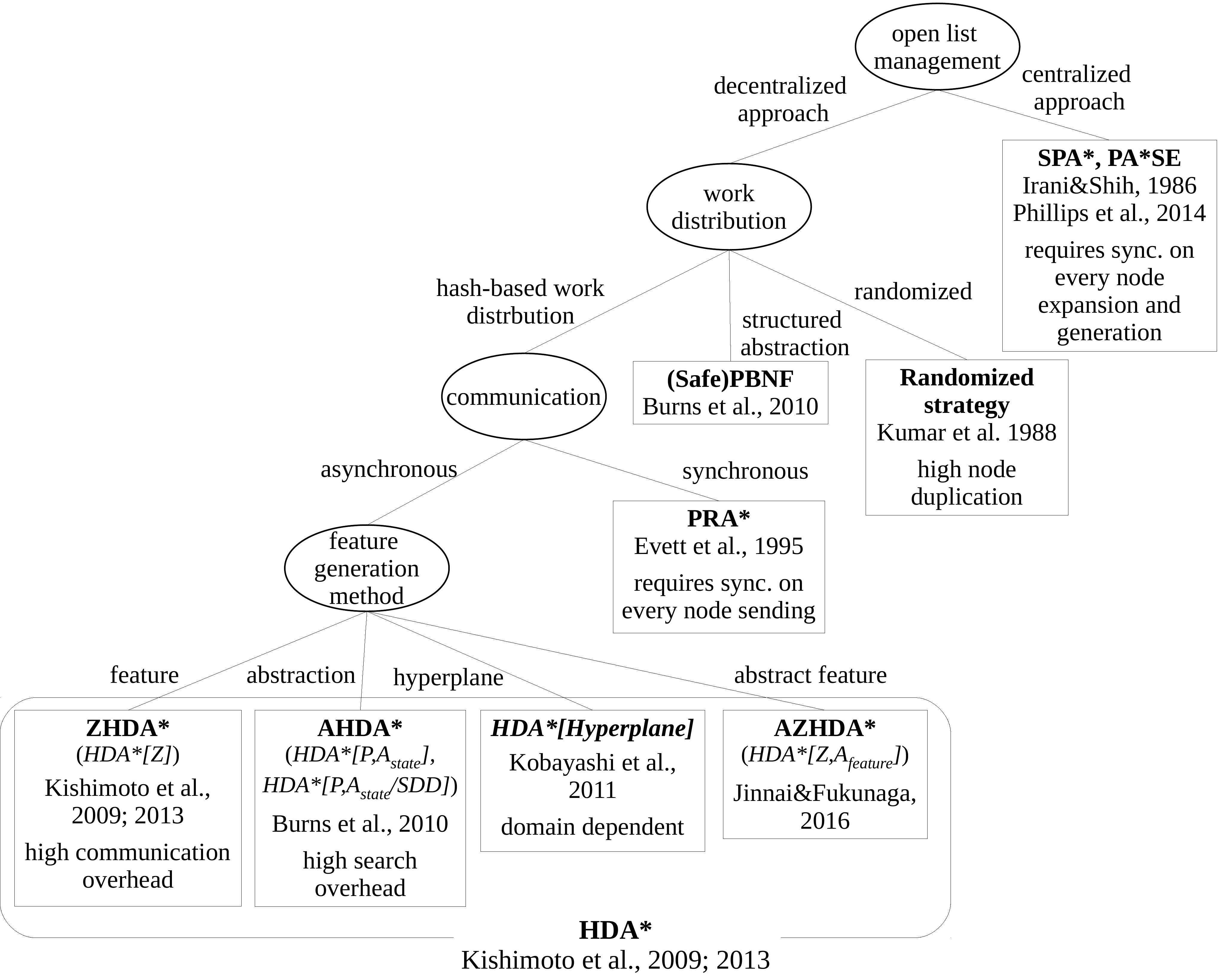}
	\caption{Classification of parallel best-first searche algorithms.}
	\label{eleven:fig:parallel-best-first}
\end{figure}

The key to achieving good speedup in parallel search is minimizing such overheads.
This is often a difficult task, in part because the overheads are interdependent.
For example, reducing search overhead usually increases synchronization and communication
overhead.
Figure \ref{eleven:fig:parallel-best-first} presents a visual classification of these approaches, which summarizes the survey of approaches in the next several sections.

\subsection{Centralized Parallel A*}

Algorithms such as breadth-first or best-first search
(including A*) use an open list which stores the set of states that have been generated but not yet expanded.
In an early study, Kumar, Ramesh, and Rao \cite{Kumar:1988}
identified two broad approaches to parallelizing best-first search, based on how the usage and maintenance of the open list was parallelized.
We first survey centralized approaches to parallel A*.

\begin{algorithm}
	
	Initialize \sharedopen to $\{s_0\}$\;
   Initialize Lock $l_o,l_i$\;
	Initialize $incumbent.cost = \infty$\;
	In parallel, on each thread, execute 5-32\;
    \While {TerminateDetection()} {
    	\If {\sharedopen $ = \emptyset$ \textbf{or} 
Smallest $f(n)$ value of $n \in $ \sharedopen $\geq incumbent.cost$} {
			Continue\;
	 	}
       AcquireLock($l_o$)\;
	    Get and remove from \sharedopen a node $n$ with a smallest $f(n)$\;
       ReleaseLock($l_o$)\;
	    Add $n$ to \sharedclosed\;
	    \If {$n$ is a goal node} {
		   AcquireLock($l_i$)\;
	      \If {path cost from $s_0$ to $n < incumbent.cost$}{
				$incumbent$ = path from $s_0$ to $n$\;
				$incumbent.cost$ = path cost from $s_0$ to $n$\;
			}
		   ReleaseLock($l_i$)\;
	    }
	    \For {every successor $n'$ of $n$} {
	    	$g_1 = g(n) + c(n,n')$\;
	    	\If {$n' \in$ \sharedclosed} {
	    		\If {$g_1 < g(n')$} {
	    			Remove $n'$ from \sharedclosed and add it to \sharedopen\;
	    		} \Else {
	    		    Continue\;
		    	}
	    	} \Else {
		    	\If {$n' \notin$ \sharedopen} {
		    		Add $n'$ to \sharedopen\;
		    	}
		    	\ElseIf {$g_1 \geq g(n')$} {
		    		Continue\;
		    	}
		    }
   			Set $g(n') = g_1$\;
   			Set $f(n') = g(n') + h(n')$\;
   			Set $\mbox{parent}(n') = n$\;
	    }
    }
	\If {$incumbent.cost = \infty$} {
	   Return failure (no path exists)\;
	} \Else {
		Return solution path from $s_0$ to $n$\;
	}
	
	\caption{Simple Parallel A* (SPA*)}
	\label{eleven:fig:spastar-pseudocode}
\end{algorithm}

The most straightforward way to parallelize A* on a shared-memory, multi-core machine is Simple Parallel A* (SPA*) \cite{iranis86}, shown in Algorithm \ref{eleven:fig:spastar-pseudocode}.
\fromaij{}
In SPA*, a single open list is
shared among all processors. Each processor expands one of the current
best nodes from the globally shared open list, and generates and evaluates its
children. This centralized approach introduces very little or no 
search overhead, 
and no load balancing among processors is necessary.
Node re-expansions are possible in SPA* because (as with most other parallel A* variants) SPA* does not guarantee that a state has an optimal $g$-value when expanded.
SPA* is especially simple to implement in a
shared-memory architecture by using a shared data structure for the
open list and closed list.
\fromjfjournal{}
However, concurrent access to the shared open list becomes a bottleneck, even if lock-free data structures are used  \cite{burnslrz10} --  in fact, for problems with fast node generation rates, 
SPA* exhibits runtimes that are {\em slower than single-threaded A*} \cite{burnslrz10}.
Thus, the scalability of the centralized approach is limited unless the time required to expand each node is extremely expensive (if the node expansion rate is slow enough, then concurrent access to the open list will not be a bottleneck).


\fromaij{relatedwork}
Vidal et al.~\cite{VBH:socs2010} propose Parallel K-Best First Search, a multi-core version of the 
K-BFS algorithm~\cite{Felner03kbfs:k-best-first}, a satisficing (non-admissible) best-first search variant.
Parallel KBFS is a centralized best-first search strategy, enhanced by the  use of more threads than the number of physical cores, which improves performance on hard problems by exploiting search diversification effects. This is further improved using a restart strategy.
 They show that good scaling behavior can be obtained on a 4-core machine.

\fromjfjournal{}
Phillips et al. have proposed PA*SE, a mechanism for reducing node re-expansions in SPA* \citeyear{PhillipsLK14} that only expands nodes when their $g$-values are optimal, ensuring that nodes are not re-expanded. 

\subsection{Decentralized Parallel A*}

\fromjfjournal{}
As described above, SPA* suffers from severe synchronization overhead due to the need to constantly access shared open/closed lists.

\begin{algorithm}
	
	Initialize $OPEN_p$ for each thread $p$\;
	Initialize $incumbent.cost = \infty$\;
	Add $s_0$ to $OPEN_{ComputeRecipient(s_0)}$\; 
	In parallel, on each thread $p$, execute 5-31\;
	 \While {TerminateDetection()} {
		 \While {$BUFFER_p \neq \emptyset$} {
		 	Get and remove from $BUFFER_p$ a triplet $(n', g_1, n)$\;
			\If {$n' \in CLOSED_p$} {
	   	 		\If {$g_1 < g(n')$} {
	    			Remove $n'$ from $CLOSED_p$ and add it to $OPEN_p$\;
	    		} \Else {
					Continue\;
				}
			}\Else {
		    	\If {$n' \notin OPEN_p$} {
		    		Add $n'$ to $OPEN_p$\;
		    	}
		    	\ElseIf {$g_1 \geq g(n')$} {
		    		Continue\;
		    	}
			}
   			Set $g(n') = g_1$\;
   			Set $f(n') = g(n') + h(n')$\;
   			Set $\mbox{parent}(n') = n$\;
		 }

	 	 \If {$OPEN_p = \emptyset$ \textbf{or} Smallest $f(n)$ value of $n \in OPEN_p \geq incumbent.cost$} {
	 	 	Continue\;
		 }
	    Get and remove from $OPEN_p$ a node $n$ with a smallest $f(n)$\;
	    Add $n$ to $CLOSED_p$\;
	    \If {$n$ is a goal node} {
	      \If {path cost from $s_0$ to $n < incumbent.cost$} {
				$incumbent$ = path from $s_0$ to $n$\;
				$incumbent.cost$ = path cost from $s_0$ to $n$\;
			}
	    }
	    \For {every successor $n'$ of $n$} {
			Set $g_1 = g(n) + c(n, n')$\;
			Add $(n', g_1, n)$ to $BUFFER_{ComputeRecipient(n)}$\;
	    }
   }
	\If {$incumbent.cost = \infty$} {
	   Return failure (no path exists)\;
	} \Else {
		Return solution path from $s_0$ to $n$\;
	}

	\caption{Decentralized A* with Local OPEN/CLOSED lists}
	\label{eleven:fig:decentralized-pseudocode}
\end{algorithm}

In contrast, in a {\em decentralized} approach to parallel best-first
search, shown in Algorithm \ref{eleven:fig:decentralized-pseudocode}, each processor has its own open list.
Initially, the root processor generates and distributes
some search nodes among the available processors. Then, each processor
starts to locally run best-first search using its local open
list (as well as a closed list, in case of 
algorithms such as A*). 
%
Decentralizing the open list eliminates the concurrency overhead associated with a 
shared, centralized open list, but load balancing becomes necessary.

Kumar, Ramesh and Rao
\cite{kumar1988parallel}, as well as Karp and Zhang \cite{karpz88,karpz93}
proposed a random work allocation strategy, where newly generated states are sent to
random processors, i.e., in the decentralized algorithm schema in Figure \ref{eleven:fig:decentralized-pseudocode}, line 31, $ComputeRecipient(n')$ simply returns a random processor ID.
In parallel architectures with non-uniform communication costs, a straightforward variant of
this randomized strategy is to send states to a random neighboring processor (with low communication cost) to avoid the cost of sending to an arbitrary processor (cf., \cite{DuttM94}).

\fromjfjournal{}
The problem with these randomized  strategies is that duplicate nodes are not detected unless they are fortuitously sent to the same processor, which can result in a tremendous amount of search overhead due to nodes that are  redundantly expanded by multiple processors.
In many search applications, including domain-independent planning,
the search space is a graph rather than a tree, and there are multiple paths to 
the same state. 
In sequential search, duplicates can be detected and pruned
by using a closed list (e.g., hash table) or other
duplicate detection techniques (e.g., \cite{Korf:2000,Zhou:2006}).
Efficient duplicate detection is critical for performance,
both in serial and parallel search algorithms, and 
can potentially eliminate vast amounts of redundant work.

In parallel search, duplicate state detection 
incurs several overheads,
depending on the algorithm and the machine environment.
For instance, in a shared-memory environment, 
many approaches, including work stealing,
need to carefully manage locks on the shared open and closed lists.





\subsubsection{Termination Detection in Decentralized Parallel Search}
\label{eleven:sec:termination-detection}
In a decentralized parallel A*, when a solution is discovered,
there is no guarantee at that time that the solution is optimal~\cite{Kumar:1988}.
When a processor discovers a locally optimal solution, the processor broadcasts its cost.
The search cannot terminate until all processors have proved that there is no
solution with a better cost.
In order to correctly terminate a decentralized parallel A*, it is not sufficient to check the local open list at every processor.
We must also ensure that there is no message en route to some processor that could lead to a better solution.
Various algorithms to handle termination exist.
A commonly used method is by Mattern \cite{Mattern:87}.

Mattern's method is based on counting sent messages and received messages.
If all processors were able to count simultaneously, 
it would be trivial to detect whether a message is still en route.
However, in reality, different processors $P_i$ will report their sent and received counters,
$S(t_i)$ and $R(t_i)$, at different times $t_i$.
To handle this, Mattern introduces a basic method where the counters are reported in two different waves.
Let $R^* = \sum_i R(t_i)$ be the accumulated received counter at the end of the first wave,
and $S'^* = \sum_i S(t'_i)$ be the accumulated sent counter at the end of the second wave.
Mattern proved that if $S'^* = R^*$, then the termination condition holds 
(i.e., there are no messages en route that can lead to a better solution).

Mattern's time algorithm is a variation of this basic method that allows checking the termination 
condition in only one wave.
Each work message (i.e., containing search states to be processed) has a time stamp,
which can be implemented as a clock counter maintained locally by each processor.
Every time a new termination check is started, the 
initiating processor increments its clock counter and sends a \emph{control} message to another processor,
starting a chain of control messages that will visit all processors and return to the first one.
When receiving a control message, a processor updates its clock counter $C$ to 
$\max (C, T)$, where $T$ is the maximum
clock value among processors visited so far. 
If a processor contains a received message $m$ with a time stamp $t_m \geq T$,
then the termination check fails.
Obviously, if, at the end of the chain of messages, the accumulated sent and received counters
differ, then the termination check fails as well.

%% file: hash-based-dec-astar.tex
\section{Hash-Based Decentralized A*}
\label{eleven:sec:hash-based-decentralized}
An approach to decentralized A* which cleanly addresses both load balancing and duplicate detection assigns a unique owner processor to each search node according to a hash function. 
That is, in Figure \ref{eleven:fig:decentralized-pseudocode}, line 31, $ComputeRecipient(n')$ is implemented by hashing, i.e., $\mathit{ComputeRecipient(n')} = \mathit{hash(n')} \thinspace \mathit{mod} \thinspace \mathit{numprocessors}$.
This maps each state to exactly one processor which ``owns'' the state.
If the hash keys are distributed uniformly among the processors, 
and the time to process each state is the same, then
load balancing is achieved.
Furthermore, duplicate detection is performed by the ``owner'' state 
-- states that are already in the local OPEN/CLOSED lists are duplicates, and by definition, nodes can never be expanded by a non-owner processor.

The idea of hash-based work distribution for parallel best-first search was first used in PRA* by Evett et al. \cite{Evett:1995}, a limited-memory best-first search algorithm for a massively parallel SIMD machine (see Section \ref{eleven:sec:pra}).
It was then used in a parallelization of SEQ\_A*, a variant of A* that performs partial expansion of states, on a hypercube by Mahapatra and Dutt \cite{MahapatraD97}, who called the technique Global Hashing (GOHA). However, the hash-based work distribution mechanism itself was not studied deeply by either Evett et al. or  Mahapatra and Dutt, as their work encompassed significantly more than this work distribution mechanism\footnote{PRA* has a sophisticated node retraction mechanism which allows more nodes to be searched in a limited amount of memory than A*, and GOHA was treated as a baseline for LOHA\&QE, a more complex mechanism which decouples duplicate checking and load balancing and also applies a more localized hash function.}
Transposition-Table-Driven Work Scheduling  
(TDS) \cite{romein1999transposition} is a distributed-memory, parallel IDA* with hash-based work distribution (see Section \ref{eleven:sec:tds}). 
Kishimoto, Fukunaga, and Botea reopened investigation into hash-based work distribution for A* by implementing HDA*, a straightforward application of hash-based work distribution to A*, showing that it scaled quite well on both multi-core machines and large-scale clusters \cite{kishimotofb09,kishimotofb13}. 
The key to achieving good parallel speedups in hash-based work distribution is the hash function. While PRA* left the hash function undefined in the paper and GOHA used a multiplicative hash function (see Section \ref{eleven:sec:goha-hash}), HDA* used the Zobrist hash function \cite{Zobrist1970}. Unfortunately, the early work on HDA* did not 
quantitatively evaluate the effect of the choice of hash function, resulting in some misleading results in later work using implementations of HDA* that did not use a hash function which was as effective as the Zobrist function.
Recently, Jinnai and Fukunaga compared hash distribution functions that have been used in the literature, showing that 
the Zobrist hash function as well as Abstract Zobrist hashing, an improved version of the Zobrist function, significantly outperforms other hash functions which have been used in the literature \cite{jinnai2016structured}.
Further details on hash functions as well as an experimental comparison are in Section \ref{eleven:sec:hash-functions}.

\subsection{Hash Distributed A*}
\label{eleven:sec:hda*}
\index{hash distributed A* (HDA*)}
We now describe details of Hash Distributed A* (HDA*), a simple, decentralized
parallelization of A* using hash-based work distribution.
In HDA* the closed and open lists are implemented as a distributed
data structure, where each processor ``owns'' a partition of the entire search space.
The local open and closed lists for processor $P$ are denoted $Open_P$ and $Closed_P$.
The partitioning is done by hashing the state, as described below.

HDA* starts by expanding the initial state
at the root processor.
Then, each processor $P$ executes the following loop until an optimal solution is found: 
\begin{enumerate}
\item
First, $P$ checks whether one or more
new states have been received in its message queue. If so, $P$
checks for each new state $s$ in $Closed_P$, in order to
determine whether $s$ is a duplicate, or whether it should be
inserted in $Open_P$.\footnote{Even if the heuristic function \cite{Helmert:07} is consistent,
parallel A* search may sometimes have to reopen a state saved in the closed list.
For example, $P$ may receive many identical states with various priorities 
from different processors and these states may reach $P$ in any order.}

\item
If the message queue is empty, then $P$ selects a highest priority
state from $Open_P$ and expands it, resulting in newly generated states.
For each newly generated state $s$, a hash key $K(s)$ is computed based on the state representation, and the re$K(s)$ and $s$
is sent to the processor that owns $K(s)$. This send is asynchronous and non-blocking. $P$ continues its computation without waiting for a reply from the destination.
\end{enumerate}

In a straightforward implementation of hash-based work
distribution on a shared-memory machine, each thread
owns a local open/closed list implemented in shared memory,
and when a state $s$ is assigned to some thread, the writer thread obtains
a lock on the target shared memory, writes $s$, then releases
the lock.  Note that whenever a thread $P$ ``sends'' a state $s$ to
a destination $dest(s)$, then $P$ must wait until the lock for
the shared open list (or message queue) for $dest(s)$ is available and not
locked by any other thread.  This results in significant
synchronization overhead -- for example, it was observed in
\cite{Burns:09} that a straightforward implementation of PRA*
exhibited extremely poor performance on the Grid search problem, 
and multi-core performance for up to 8 cores was consistently {\em slower}
than sequential A*.
While it is possible to speed up locking operations by using, for
example, highly optimized implementations of lock operations in inline
assembly language, 
the performance degradation due to synchronization 
remains a considerable problem.

In contrast, the open/closed lists in HDA* are not explicitly shared among the processors. 
Thus, even in a multi-core environment where it is possible to share memory, all communications are done 
between separate MPI processes using non-blocking send/receive operations (e.g. {\tt MPI\_Bsend} and {\tt MPI\_Iprobe}).
and relies on highly optimized message buffers
implemented in  MPI.

Every state must be sent from the processor where it is generated to its ``owner'' processor.
In their work with transposition-table-driven scheduling for parallel IDA*, 
Romein et al. \cite{Romein:99} showed that this communication overhead could be overcome by packing multiple 
states with the same destination into a single message.
HDA* uses this state-packing strategy to reduce the number of messages.
The relationship between performance and message sizes depends on several factors such as network configurations, the number of CPU cores, and CPU speed.
In \cite{kishimotofb09,kishimotofb13},  100 states are packed into each message on a commodity cluster using
more than 16 CPU cores and a HPC cluster, while 10 states are packed on the commodity cluster
using fewer than 16 cores. 

%% file: dec-search-str-part.tex
\section{Decentralized Search Using Structure-based Search Space Partitioning)}
\label{eleven:sec:structure-based}

An alternate approach for load balancing is based on {\it structured abstraction}. 
Given a state space graph and a projection function, an abstract state graph is (implicitly) generated by projecting states from the original state space graph into abstract nodes.
In many domains, a projection function can be derived by ignoring some features in the original state space.
For example, an abstract space for the sliding-tile puzzle domain can be created by projecting all nodes with the blank tile at position $b$ to the same abstract state.
While the use of abstractions as the basis for heuristic functions has a long history \cite{Pearl84}, 
the use of abstractions as a mechanism for partitioning search states originated in Structured Duplicate Detection (SDD), \index{structured duplicate detection}
an external memory search which stores explored states on disk \cite{zhou2004structured}.
In SDD, an $n$-block is defined as the set of all nodes which map to the same abstract node.
SDD uses $n$-blocks to enable duplicate detection.
For any node $n$ that belongs to  $n$-block $B$, the {\it duplicate detection scope} of $n$ is defined as the set of $n$-blocks that can possibly contain duplicates of $n$, and duplicate checks can be restricted to the duplication detection scope, thereby avoiding the need to look for a duplicate of $n$ outside this scope.
SDD exploits this property for external memory search by expanding nodes within a single $n$-block $B$ at a time and keeping the duplicate detection scope of the nodes in $B$ in RAM, avoiding costly I/O.
Parallel Structured Duplicate Detection (PSDD) \index{parallel structured duplicate detection}
is a parallel search algorithm that exploits $n$-blocks to address both  synchronization overhead and communication overhead \cite{zhou2007parallel}. Each processor is exclusively assigned to an $n$-block and its neighboring $n$-blocks (which are the duplication detection scopes). By exclusively assigning $n$-blocks with disjoint duplicate detection scopes to each processor, synchronization during duplicate detection is eliminated.
While PSDD uses disjoint duplicate detection scopes to parallelize breadth-first heuristic search \cite{zhou2006breadth}, 
Parallel Best-NBlock-First (PBNF) \cite{burnslrz10} \index{PBNF}
extends PSDD to best-first search on multi-core machines by ensuring that $n$-blocks with the best current $f$-values are assigned to processors.

Since livelock is possible in PBNF  on domains with infinite state spaces, Burns et al. proposed  SafePBNF, a livelock-free version of PBNF \citeyear{burnslrz10}. 
Burns et al. \citeyear{burnslrz10} also proposed AHDA*, a variant of HDA* using an abstraction-based node distribution function. AHDA* is described below in Section \ref{eleven:sec:ahda}.

%% file: hash-functions-hdastar.tex
\section{Hash Functions for Hash-Based Decentralized Work Distribution}
\label{eleven:sec:hash-functions}

\fromjfjournal{background}

The performance of hash-based decentralized A* algorithms in Section \ref{eleven:sec:hash-based-decentralized} depends
entirely on the characteristics of the hash function.
However, early work on hash-based decentralized A* did not present empirical evaluation of candidate hash functions, and the importance of the choice of hash function was not fully understood or appreciated. Recent work has investigated the performance characteristics and tradeoffs among various hashing strategies, resulting in a significantly better understanding of previous hashing strategies, as well as new hashing strategies that combine previous methods in order to obtain superior  performance \cite{jinnai2016structured,jinnai2017work}.

In this section, we first classify and review various hash functions which have been proposed for hash-based distributed A* (Sections \ref{eleven:sec:goha-hash}-\ref{eleven:sec:hwd}).
We then present an evaluation of some of the functions on the sliding-tile puzzle benchmark domain (Section \ref{eleven:sec:hash-comparison}).
Next, we review fully automated, domain-independent methods for deriving hash functions (Section \ref{eleven:sec:domain-independent}).
Finally, we briefly review work on hash-based work distribution in the related field of model checking (Section \ref{eleven:sec:hash-based-model-checking}).

\subsection{Multiplicative Hashing}
\label{eleven:sec:goha-hash}

The multiplication method $H(\kappa)$ is a widely used hashing method that has been observed to hash a random key to $P$ slots with almost equal likelihood \cite{cormen01}.
Multiplicative hashing $M(s)$ uses this function to achieve good load balancing of nodes among processors \cite{MahapatraD97}:

\begin{align}
M(s) &= H(\kappa(s)), \\
H(\kappa) &= \lfloor p (\kappa \cdot A - \lfloor \kappa \cdot A \rfloor)\rfloor,
\end{align}

\noindent where $\kappa(s)$ is a key derived from the state $s$, $p$ is the number of processors, and $A$ is a parameter in the range  $[0,1)$. Typically $A = (\sqrt{5} - 1) / 2$ (the golden ratio) is used since the hash function is known to work well with this value of $A$ \cite{knuth73}. 
As $H(\kappa)$ achieves almost perfect load balance for {\it random} $\kappa$ keys, designing $\kappa(s)$ so that it appears to be random to state $s$ is important to its performance. However, designing such a $\kappa(s)$ for a given domain is a non-trivial problem. 



\subsection{Zobrist Hashing} 

\label{eleven:sec:zhda}
Since the work distribution in HDA* is completely determined by a global hash function, the choice of the hash function is crucial to its performance.
Kishimoto et al. \citeyear{kishimotofb09,kishimotofb13} noted that it is  desirable to use a hash function that uniformly distributed nodes among processors, and used the Zobrist hash function \citeyear{Zobrist1970}, \index{hashing, Zobrist}
described below. 
The Zobrist hash value of a state $s$, $Z(s)$, is calculated as follows. For simplicity, assume that $s$ is represented as an array of $n$ propositions, $s = (x_0, x_1,..., x_n)$. Let $R$ be a table containing preinitialized random bit strings: 
\newcommand{\xor}{{\mbox{\textit{xor}}}}
\begin{equation}
\label{eleven:eq:zobrist}
	Z(s) := R[x_{0}]\; \xor \; R[x_{1}]\; \xor\; \cdots\; \xor\; R[x_{n}]%
\end{equation}

In the rest of the paper, we refer to the original version of HDA* by Kishimoto et al. \cite{kishimotofb09,kishimotofb13}, which  used Zobrist hashing, as ZHDA* or \ZHDA{}. 

It is possible for two different states to have the same Zobrist hash key, although the probability of such a collision is extremely low with 64-bit keys.
Thus, when using Zobrist hashing, checking whether a state $s$ is a duplicate requires first checking whether the bucket for $hash(s)$ is nonempty, and if so, the state itself needs to be compared.
Although this is slightly slower than
comparing only the hash key, duplicate checks are guaranteed to be
correct.

\subsection{Operator-Based Zobrist hashing} 

\fromjfjournal{was part of HDA*/Zobrist description; promoting to subsection}

Zobrist hashing seeks to distribute nodes uniformly among all processors, without any consideration of the neighborhood structure of the search space graph. 
As a consequence,
communication overhead is high.
Assume an ideal implementation that assigns nodes uniformly among threads.
Every generated node is sent to another thread with probability $1-\frac{1}{\#threads}$.
Therefore, with 16 threads, $>90\%$ of the nodes are sent to other threads,
so
communication costs are incurred for the vast majority of node generations.

Operator-based Zobrist hashing (OZHDA*) \cite{jinnai2016automated}  \index{hashing, operator-based Zobrist}
partially addresses this problem by manipulating the random bit strings in $R$, the  table used to compute Zobrist hash values, such that for some selected states $S$, there are some operators $A(s)$ for  $s \in S$ such that the successors of $s$ that are generated when $a \in A(s)$
is applied to $s$ are guaranteed to have the same Zobrist hash value as $s$, which ensures that they are assigned to the same processor as $s$.
Jinnai and Fukunaga \citeauthor{jinnai2016automated} showed that OZHDA* significantly reduces  communication overhead compared to Zobrist hashing \citeyear{jinnai2016automated}. However, this may result in increased search overhead compared to \ZHDA, and it is not clear whether the extent of the increased search overhead in OZHDA* could be predicted \emph{a priori}. 



\subsection{Abstraction } 
\label{eleven:sec:ahda}


In order to minimize communication overhead in HDA*,  
Burns et al. \citeyear{burnslrz10} proposed AHDA*, which uses \emph{abstraction} based node assignment. \index{hashing, abstraction-based}  
AHDA* applies the state-space partitioning technique used in PBNF \cite{burnslrz10} and PSDD \cite{zhou2007parallel}.
Abstraction uses the abstraction strategy to project nodes in the state space to \emph{abstract states}. A hash-based work distribution function can then be applied to the projected state. 
The AHDA* implementation by Burns et al. \citeyear{burnslrz10} 
assigns abstract states to processors using a perfect hashing and a modulus operator.

Thus, nodes that are projected to the same abstract state are assigned to the same thread. 
If the abstraction function is defined so that children of node $n$ are usually in the same abstract state as $n$, then communication overhead is minimized.
The drawback of this method is that it focuses solely on minimizing communication overhead, and there is no mechanism for equalizing load balance, which can lead to high search overhead.


HDA* with abstraction can be characterized by two parameters to decide its behavior -- a hashing strategy and an abstraction strategy.
Burns et al. \citeyear{burnslrz10} implemented the hashing strategy using a perfect hashing and a modulus operator, and an abstraction strategy following the construction for SDD \cite{zhou2006} (for domain-independent planning), or a hand-crafted abstraction (for the sliding-tile puzzle and grid path-finding domains). 



Jinnai and Fukunaga  showed that AHDA* with a static $N_{max}$ threshold performed poorly for a benchmark set with varying difficulty because a fixed size abstract graph results in very poor load balance, and proposed Dynamic AHDA* (DAHDA*), which dynamically sets the size of the abstract graph according to the number of features (the state space size is exponential in the number of features) \cite{jinnai2016automated}.

\subsection{Abstract Zobrist Hashing}
\label{eleven:sec:azh}
\fromjfjournal{AZHsection}
Both search and communication overheads have a significant impact on the performance of HDA*, and methods that only address one of these overheads are insufficient.
ZHDA*, which uses  Zobrist hashing, assigns nodes uniformly to processors, achieving near-perfect load balance, but at the cost of incurring communication costs on almost all state generations.
On the other hand, abstraction-based methods such as PBNF and AHDA* significantly reduce communication overhead by trying to keep generated states at the same processor as where they were generated, but this results in significant search overhead because all of the productive search may be performed at one node, while all other nodes are searching unproductive nodes that would not be expanded by A*.
Thus, we need a more balanced approach that simultaneously addresses both search and communication overheads.

{\it Abstract Zobrist hashing} (AZH) is a hybrid hashing strategy which  \index{hashing, abstract Zobrist}
augments the Zobrist hashing framework with the idea of projection from abstraction, incorporating the strengths of both methods.
The AZH value of a state, $AZ(s)$ is:
\begin{equation}
\label{eleven:eq:sz}
	AZ(s) := R[A(x_{0})] \; \xor \; R[A(x_{1})]\; \xor \; \cdots \; \xor \; R[A(x_{n})]
\end{equation}
where $A$ is a {\it feature projection function}, a many-to-one mapping from each raw feature to an {\it abstract feature}, and $R$ is a precomputed table 
for each abstract feature.

\begin{figure}[htb] 
	\centering
	\subfloat[Zobrist hashing]{{\includegraphics[width=0.75\linewidth]{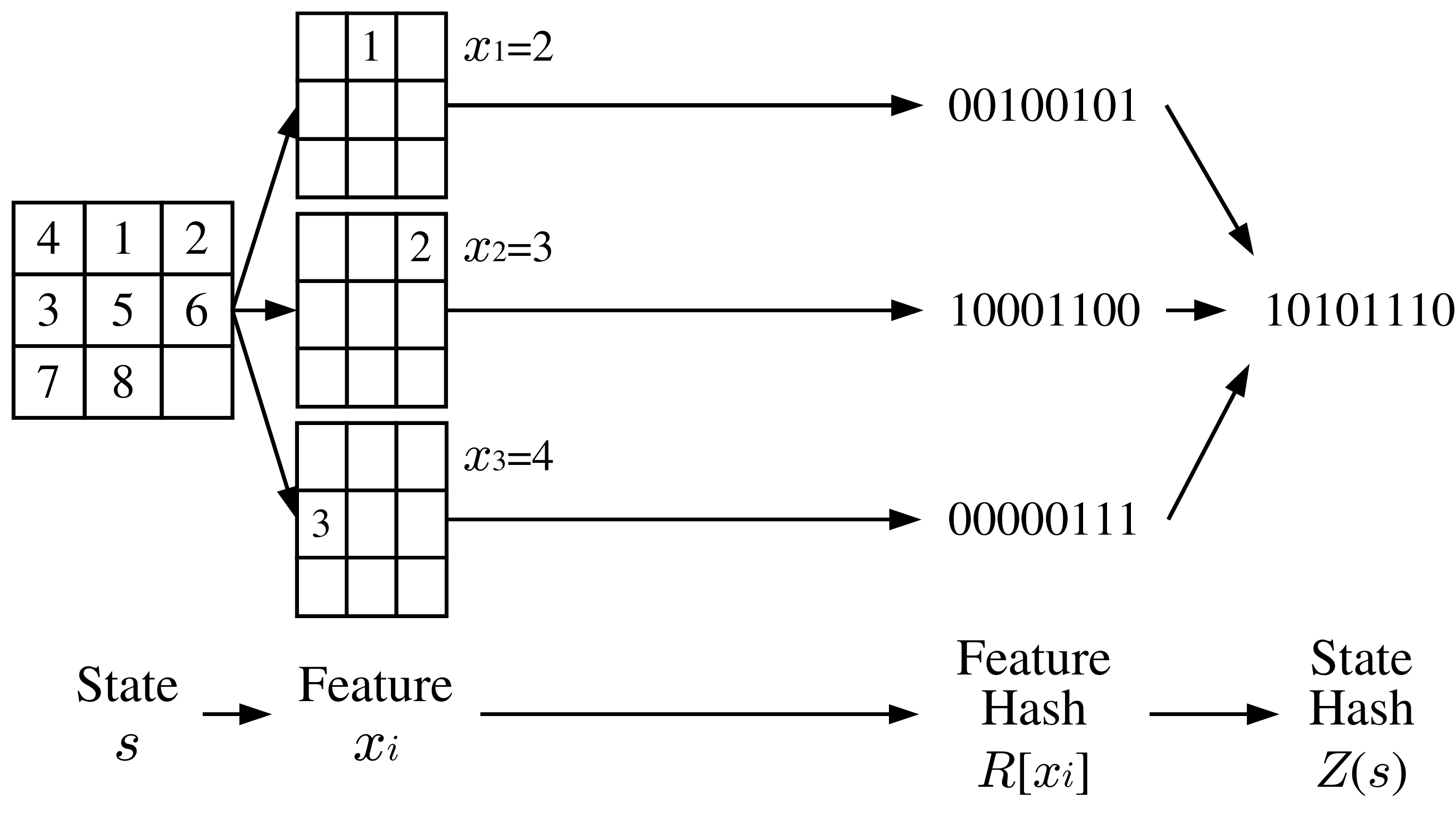}}}

	\subfloat[Abstract Zobrist hashing]{{\includegraphics[width=0.75\linewidth]{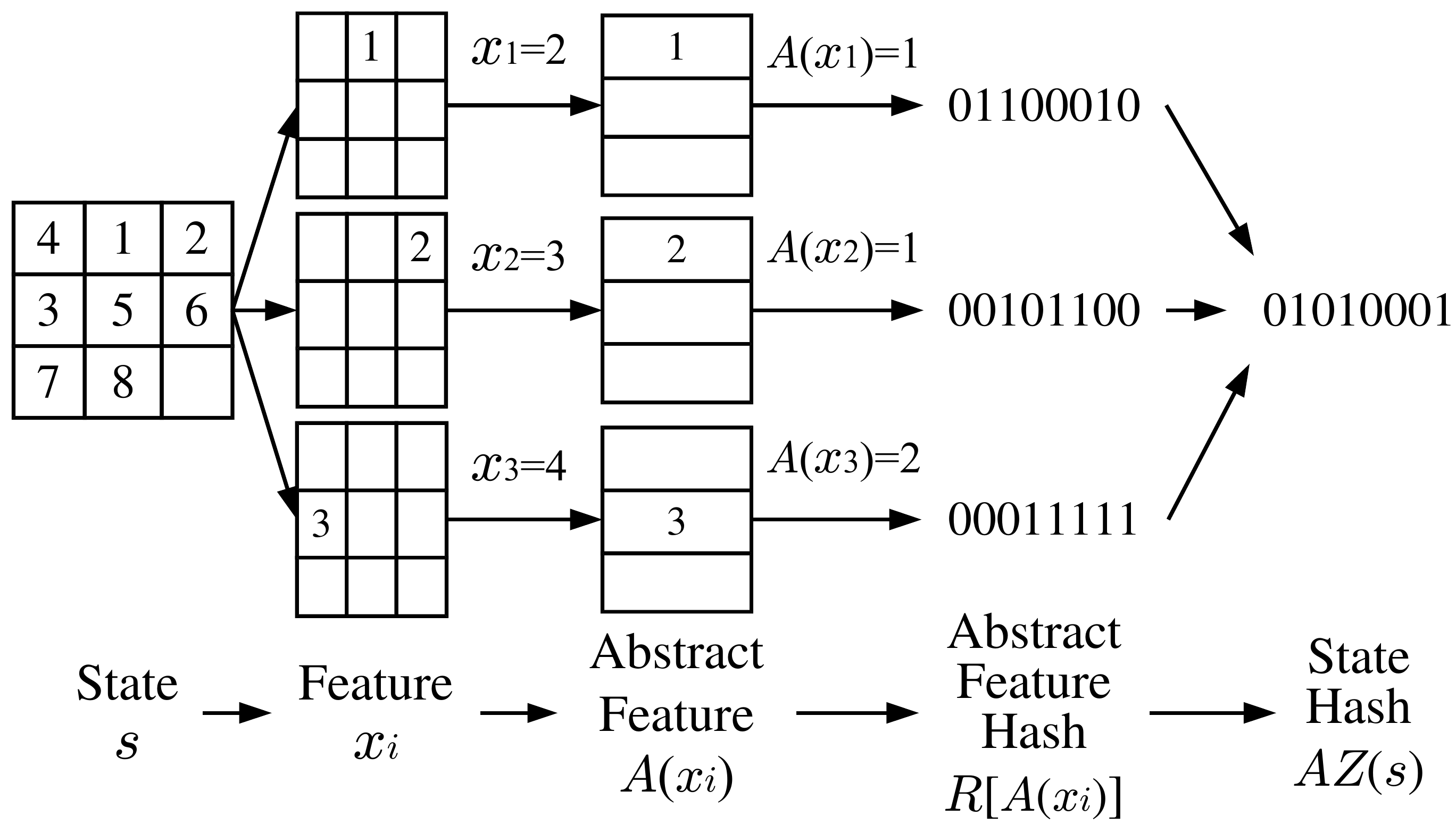}}}
	\caption{\fromjfjournal{Fig7} \small Calculation of abstract Zobrist hash (AZH) value $AZ(s)$ for the 8-puzzle: 
State 
$s = (x_1, x_2,..., x_8)$, where $x_i = 1, 2,..., 9$ ($x_i = j$ means tile $i$ is placed at position $j$).
The Zobrist hash value of $s$ is the result of xor'ing a preinitialized random bit vector $R[x_{i}]$ for each feature (tile) $x_i$. AZH incorporates an additional step which projects features to abstract features (for each feature $x_i$, look up $R[A(x_{i})]$ instead of $R[x_{i}]$).}
	\label{eleven:fig:hash-calculation}
\end{figure}

Thus, AZH is a 2-level, hierarchical hash, where 
raw features are first projected to abstract features, and Zobrist hashing is applied to the abstract features.
In other words, we project state $s$ to an abstract state $s' = (${\small $A(x_{0}), A(x_{1}),...,A(x_{n})$}$)$, and $AZ(s) = Z(s')$.
Figure \ref{eleven:fig:hash-calculation} illustrates the computation of the AZH value for an 8-puzzle state.

AZH seeks to combine the advantages of both abstraction and Zobrist hashing.
Communication overhead is minimized by building abstract features that share the same hash value (abstract features are analogous to how abstraction projects state to abstract states), and load balance is achieved by applying Zobrist hashing to the abstract features of each state. 

Compared to Zobrist hashing, AZH incurs less CO due to abstract feature-based hashing.
While Zobrist hashing assigns a hash value to each node independently, AZH assigns the same hash value to all nodes that share the same abstract features for all features, reducing the number of node transfers. 
Also, in contrast to abstraction-based node assignment, which minimizes communications but does not optimize load balance and search overhead,
AZH seeks good load balance, because the node assignment considers
all features in the state, rather than just a subset.

AZH is simple to implement, requiring only an additional projection per feature compared to Zobrist hashing, and we can precompute this projection at initialization. Thus, there is no additional runtime overhead per node during the search. The projection function $A(x)$ can be either hand-crafted or automatatically generated.

\subsection{Hyperplane Work Distribution}
\label{eleven:sec:hwd}

HDA* suffers significantly from increased search overhead in the multiple sequence alignment (MSA) domain whose search space is a directed acyclic graph
with non-uniform edge costs \cite{KobayashiKW2011}.  
The increased search overhead is caused by reopening the nodes in the closed list to ensure solution optimality.  
Even with a consistent heuristic, HDA* may need to reopen a node, 
because HDA* selects the best node in its local open list, which is not necessarily the globally best node. 
On the other hand, A* with the consistent heuristic never reopens the nodes in the closed list.

\begin{figure}[htb] 
 \begin{center}
  \includegraphics[scale=1]{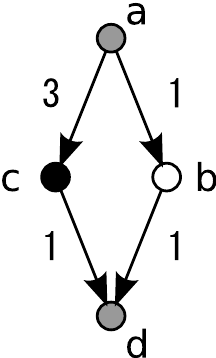}
 \end{center}
 \caption{An example from \cite{KobayashiKW2011}, showing how  HDA* may expand nodes in a non-optimal order, resulting in duplicate search effort }
 \label{eleven:figure:missorder}
\end{figure}

Figure \ref{eleven:figure:missorder} 
illustrates an example of HDA*'s drawback. 
Assume that $P_1$ owns states $a$, $c$, and $d$, and $P_2$ owns state $b$. 
$P_1$ is likely to expand $d$ via path $a \rightarrow c \rightarrow d$, since $P_1$ does not send $a$, $c$, and $d$ to $P_2$, while $b$ needs to be sent to $P_2$. 
Assume that $P_1$ saves $d$ in the closed list with $g(d)= 1+3=4$ and expands $d$, then receives $d$ from $P_2$ via $a \rightarrow b \rightarrow d$ with $g(d) = 1 + 1 = 2$, and saves $d$ in the open list. Then, when choosing $d$ for expansion, $P_1$ needs to regenerate the successors of $d$. 

\def\round{\mathrm{round}}
\def\Zobrist{\mathrm{Z}}
\def\Plane{\mathrm{Plane}}
\def\Succ{\mathrm{Succ}}
\
In MSA with $n$ sequences, a state can be represented by a location
$\vec{x} = (x_1,x_2, \cdots, x_n)$ in the $n$-dimensional grid,
where $x_i$ is an integer ($0 \leq x_i \leq l_i$) and $l_i$ is the length of the $i$-th sequence. 
Based on the hyperplane defined in the structural regularity in the MSA search space,  
the hyperplane work distribution (HWD) strategy attempts to limit the owners of successors to some processors.  \index{hashing, hyperplane work distribution}
In HWD, the owner of state $\vec{x}$ is defined as:

\[
{\small
 \Plane(\vec{x},d)
  := \left\{
   \begin{array}{ll}
	\lfloor \frac{1}{d}\sum x_i \rfloor
	 & \bigl( d \in \{1,2,3,...\} \bigr) \nonumber \\ 
	\frac{1}{d}\sum x_i + \left( \Zobrist(\vec{x}) \mod \frac{1}{d} \right)
	 & \bigl( d \in \{\frac{1}{2},\frac{1}{3},...,\frac{1}{p}\} \bigr)
   \end{array}
  \right.
}
\]

\noindent where $p$ is the number of processors, 
$d$ is an empirically determined parameter indicating the {\it thickness} of the hyperplane, 
and $Z$ is the Zobrist function. 
Then, processor $P_i \ (0 \leq i < p)$ owns $\vec{x}$ where $i=P(\vec{x})$ and $P(\vec{x}):=\Plane(\vec{x},d) \mod p$. 

HWD's work localization scheme increases the chance of allocating generated successors to the same processor. 
The local open list of HWD orders these successors more reasonably, thus contributing to reducing the frequency of reopening the states. 
For example, if states $b$ and $c$ are allocated to the same processor and $h(b)=h(c)$ holds, the processor expands $b$ before $c$. Thus, $d$ via $a \rightarrow b \rightarrow d$ is generated first,
and $d$ via $a \rightarrow c \rightarrow d$ is successfully removed.  

Assume processor $P_i$ owns $\vec{x}$ and let $\Succ(x)$ be a set of successors of $\vec{x}$. 
Then, the following theorem indicates that HWD bounds the number of processors to
which $P_i$ sends the successors of $\vec{x}$. 

\begin{theorem} \label{eleven:theorem:childsize}
{\small 
 \[
 \# \left(\bigcup_{x~:~P(x)=i} \left\{ P(\vec{x'})~|~\vec{x'} \in \Succ(\vec{x}) \right\}\right)
 \le
 \left\lfloor \frac{n}{d}+ \max(1, \frac{1}{d}) \right\rfloor
 \]
}
\end{theorem}

There is a trade-off between load balancing and localization of the work. Choosing a good value for $d$ is important for
achieving satisfactory parallel performance (see \cite{KobayashiKW2011} for details).

LOHA \cite{MahapatraD97} distributes work with a hash function taking into account locality for the Traveling Salesperson Problem where
the search space is represented as a levelized graph. 
LOHA is similar to HWD in the sense that both approaches limit the number of destination processors to which each processor sends work. 
However, there are notable differences between LOHA and HWD in the design of the hash functions. 
LOHA does not employ the Zobrist function, 
\todo{What is the hash function LOHA uses -- the GOHA multiplicative function?}
 which plays an important role for uniformly distributing work. 
In addition, LOHA was designed for the Hypercube machine whose communication delays between subcubes are much larger than between processors inside the same subcube. 
As a result, LOHA first allocates coarse-grained work to a subcube, then splits such allocated work finely among the processors inside the subcube. 
On the other hand, HWD directly partitions fine-grained work to a restricted subset of processors, aiming to reduce search overhead incurred by reopening the states. 

Both HWD and LOHA require the search space to be levelized. Their extension to non-levelized graphs such as cost-optimal planning remains an open question. 

\subsection{Empirical Comparison of Hash Functions}
\label{eleven:sec:hash-comparison}



To illustrate the scaling behavior of the various hash functions reviewed in this section, 
We evaluated the performance of the following parallel A* algorithms on the 15-puzzle.
See \cite{jinnai2017work} for a more detailed comparisons 

\begin{itemize}
\item AZHDA*: HDA* using Abstract Zobrist hashing \cite{jinnai2016structured}

\item ZHDA*: HDA* using Zobrist hashing \cite{kishimotofb13}

\item AHDA*: HDA* using abstraction based work distribution \cite{burnslrz10}

\item SafePBNF: \cite{burnslrz10}


\item HDA*+GOHA: HDA* using multiplicative hashing, a hash function proposed in \cite{MahapatraD97}

\item Randomized strategy: nodes are sent to random cores (duplicate nodes are not guaranteed to be sent to the same core) \cite{kumar1988parallel,karpz93}

\item Simple Parallel A* (centralized, single OPEN list)  \cite{iranis86} 

\end{itemize}

This experiment was run on an Intel Xeon E5-2650 v2 2.60 GHz CPU with 128 GB RAM, using up to 16 cores. The code for the experiment (based on the code by \cite{burnslrz10}) is available at \url{https://github.com/jinnaiyuu/Parallel-Best-First-Searches}.

We solved 100 randomly generated instances using Manhattan distance heuristic. Following \cite{burnslrz10}, we implemented open list using a binary heap. The average runtime of sequential A* solving these instances was 52.3 seconds. 

\begin{figure}[htb]
	\centering
	\subfloat[15-puzzle ZHDA]{\hspace{12pt}{\includegraphics[width=0.15\linewidth]{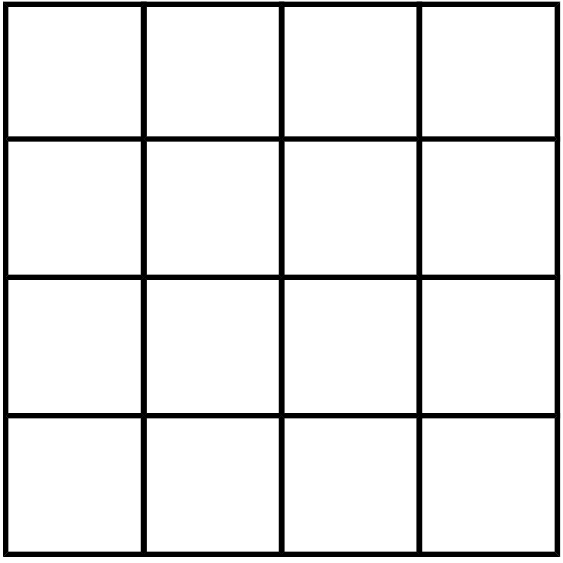}}\hspace{12pt}} \hspace{2pt}%
	\subfloat[15-puzzle AZHDA]{\hspace{12pt} \label{eleven:fig:15azh} {\includegraphics[width=0.15\linewidth]{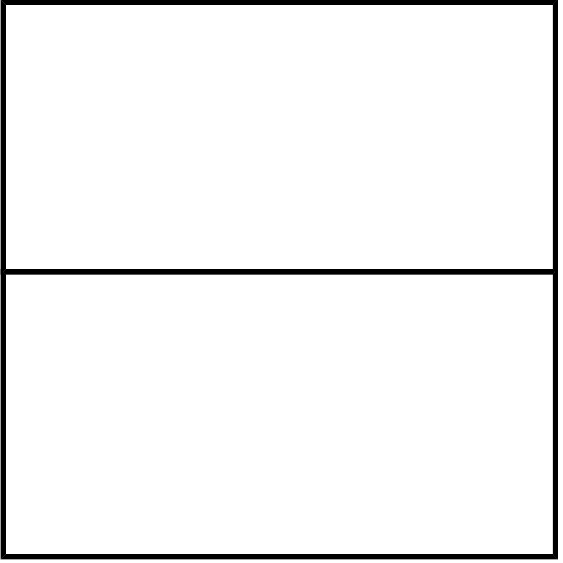} } \hspace{12pt}} \hspace{2pt}%
	\caption{The hand-crafted abstract features used by abstract Zobrist hashing for the 15-puzzle.}
	\label{eleven:fig:projection-used}
\end{figure}

The features used by Zobrist hashing in ZHDA* are the positions of each tile $i$. The projections we used for Abstract Zobrist hashing in AZHDA* are shown in Figure \ref{eleven:fig:projection-used}. The abstraction use by AHDA* and SafePBNF ignores the positions of all tiles except tiles 1, 2, and 3.
For HDA*+GOHA, we used a bit vector of the positions of the tiles for $\kappa$.

\begin{figure}[htb]
	\centering
	\includegraphics[width=1.0\linewidth]{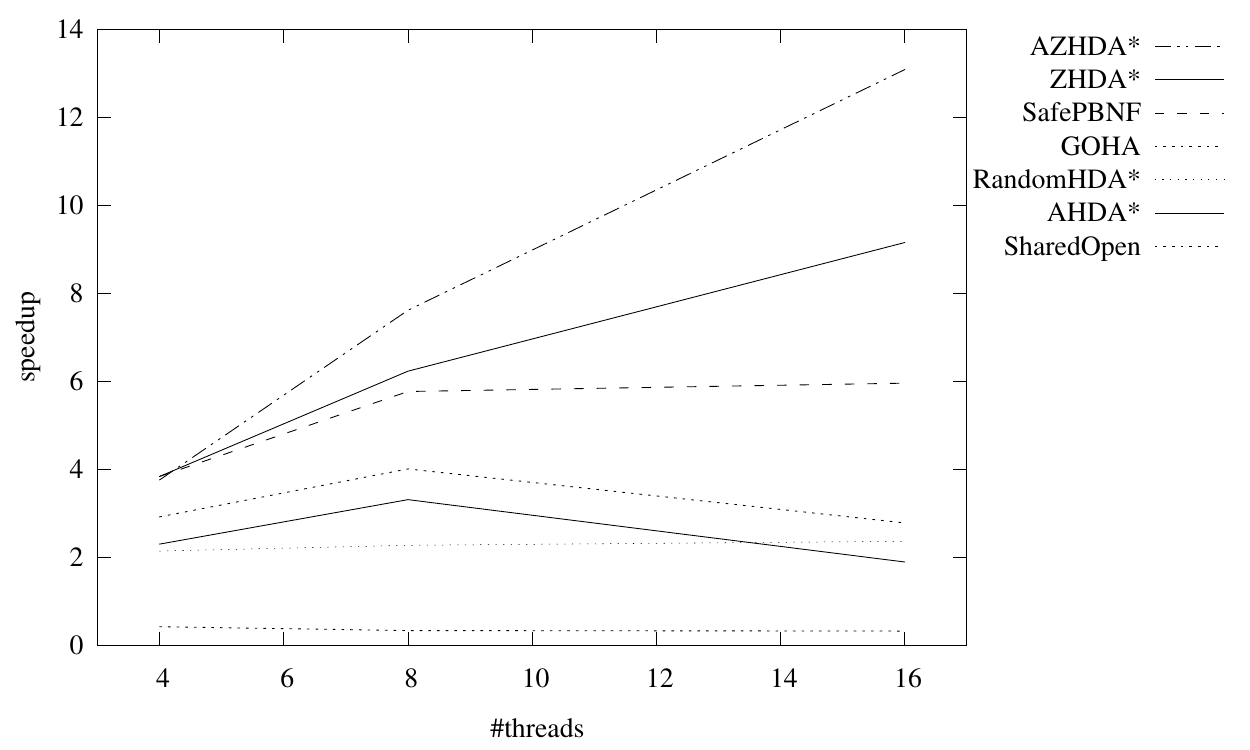}
	\caption{Comparison of speedups obtained by HDA* variants using various hashing methods}
	\label{eleven:fig:pastar-speedup-comparison}
\end{figure}

Figure \ref{eleven:fig:pastar-speedup-comparison} shows the speedup of each method.

\subsection{Domain-Independent, Automatic Generation of Hash Functions}
\label{eleven:sec:domain-independent}

The hashing methods described above are domain-independent methods that can be applied to a wide range of problems.
Although concrete implementations of hash functions for a specific problem can be hand-crafted, as in the case of the sliding-tile puzzle example above, 
it is possible to fully automate this process when a formal model of a domain (such as PDDL/SAS+ for classical planning) is available.
For example, for ZHDA*, domain-independent feature generation for classical planning problems represented in the SAS+ representation \cite{backstrom1995complexity} is straightforward \cite{kishimotofb13}.
For each possible assignment of value $k$ to variable $v_i$ in a SAS+ representation, e.g., $v_i=k$, there is a binary proposition $x_{i,k}$ (i.e., the corresponding STRIPS propositional representation).
Each such proposition $x_{i,k}$ is a feature to which a randomly generated bit string is assigned, and the Zobrist hash value of a state can be computed by xor'ing the propositions that describe the state, as in Equation \ref{eleven:eq:zobrist}.

For AHDA*, the abstract representation of the state space can be generated by ignoring some of the features (SAS+ variables) and using the rest of the features to represent the abstraction. 
Burns et al.~\citeauthor{burnslrz10} used the greedy abstraction algorithms by Zhou and Hansen \cite{zhou2006} to select the subset of features \citeyear{burnslrz10}. The greedy abstraction algorithm adds one atom group to the abstract graph at a time, choosing the atom group which minimizes the maximum out-degree of the abstract graph, until the graph size (number of abstract nodes) reaches the threshold given by a parameter. 


For AZHDA*, the feature projection function, which generates abstract features from raw features, 
plays a critical role in determining the performance of AZHDA*, because AZHDA* relies on the feature projection in order to reduce communications overhead. Methods based on the domain-transition graph are proposed in \cite{jinnai2016automated,jinnai2017work}.

\subsection{Hash-Based Work Distribution in Model Checking}
\label{eleven:sec:hash-based-model-checking}
\fromaij{relatedwork}
While this paper focuses on parallel best-first search (more specifically, parallel A*), which is applied to standard AI search domains including domain-independent planning and the sliding-tile puzzle,
distributed search, including hash-based work distribution, has also been studied extensively by the parallel model checking community.
Parallel Mur$\varphi$~\cite{SternD97,SternD01} addresses verification tasks
that involve exhaustively enumerating all reachable states in a state space,
and implements a hash-based work distribution schema where each state is 
assigned to a unique owner processor. 
Kumar and Mercer~\cite{Kumar05loadbalancing} present a load balancing technique
as an alternative to the hash-based work distribution implemented in Mur$\varphi$.
The Eddy Murphi model checker~\cite{Melatti:2009} specializes processors' tasks,
defining two threads for each processing node. The worker thread performs state processing
(e.g., state expansion), whereas the other thread handles  communication (e.g., sending and receiving states). 

Lerda and Sisto parallelized the SPIN model checker to increase the 
availability of memory resources~\cite{lerda-99}.
Similarly to hash-based distribution,
states are assigned to an owner processing node, and get expanded at their owner node.
However, instead of using a hash function to determine the owner processor,
only one state variable is taken into account.
This is done to increase the likelihood that the processor where a state is
generated is identical to the owner processor.
Holzmann and Bo\^{s}na\^{c}ki~\cite{HolzmannB07} introduce an extension
of SPIN to multi-core, shared memory machines. 
Garavel et al.~\cite{GaravelMS01} use hash-based work distribution to convert an implicitly defined
model-checking state space into an explicit file representation.
Symbolic parallel model checking has been addressed in~\cite{DBLP:conf/cav/HeymanGGS00}.

Thus, hash-based work distribution and related techniques for distributed search have been widely studied for parallel model checking.
There are several important differences between previous work in model checking and this 
\ifhandbook
chapter.
\else
paper.
\fi
First, this 
\ifhandbook
chapter
\else
paper 
\fi 
focuses on parallel A*. In model checking, there is usually no heuristic evaluation function, 
so depth-first search
and breadth-first search is used instead of best-first strategies such as A*.

Second, reachability analysis in model checking
(e.g.,~\cite{SternD97,SternD01,lerda-99,GaravelMS01}),
which involves visiting all reachable states,
does not necessarily require optimality.
Search overhead is not an issue because 
both serial and parallel solvers will expand all reachable states exactly once.
In contrast, A* specifically addresses the problem of finding an optimal path, a significant constraint which introduces the issue of search efficiency because distributed A* (including HDA*) searches many nodes with $f$-cost greater than or equal to the optimal cost, as detailed in Section~\ref{eleven:sec:parallel-overheads}; 
furthermore, node re-expansions in parallel A* can introduce search overhead. 



%

%% file: portofolios-astar.tex
\section{Parallel Portfolios using A*}
\label{eleven:sec:portfolio-astar}

An {\em algorithm portfolio} \cite{HubermanLH97} is often employed and parallelized in other domains, such as the ManySAT solver \cite{HamadiJS09} for SAT solving and ArvandHerd \cite{ValenzanoNMSS12} for satisficing planning.  
This approach runs a set of different search algorithms in parallel. Processors execute the search algorithms mostly independently, but may periodically exchange important information with others. 

A long-tailed distribution is often observed in the runtime distribution of the search algorithms \cite{GomesSCK00}. 
The algorithm portfolio attempts to exploit such search behaviors by using
a variety of algorithms that examine potentially overlapping, but different portions of
the search space. 

Dovetailing \cite{Knight93}, which is a simple version of the algorithm portfolio, performs search simultaneously with different parameter settings. 
Valenzano et al. apply parallel dovetailing \cite{ValenzanoSSBK10} to the weighted versions of IDA* \cite{korf:85a}, RBFS \cite{korf:93}, A*, PBNF \cite{burnslrz10}\footnote{The suboptimality of these weighted algorithms is is bounded by the values of the weights.}, 
as well as BULB \cite{furcy:05}, a suboptimal heuristic search. Their parallel dovetailing runs search with many different weight values without exchanging information, and terminates
when one of the algorithms returns a solution. 

In their experiments on puzzle solving, admissible heuristics were used to evaluate the performance of parallel dovetailing. 
On sequential planning, weighted A* was executed with many different weights including the weight value of $\infty$ (i.e., identical to Greedy Best-First Search (GBFS)), one admissible heuristics and two inadmissible heuristics. 
In addition, the original Fast Downward planner using multiple heuristics and GBFS is included as one of the algorithms. 

In both puzzle solving and sequential planning, the experimental results shown by Valenzano et al. \cite{ValenzanoSSBK10} indicate that parallel dovetailing often yields good speedups and solves additional problem instances. 
However, unlike other approaches described in this article, parallel dovetailing does not always return optimal solutions. 

%% file: mem-lim-astar.tex
\section{Parallel, Limited-Memory A* (Parallel IDA*, TDS, PRA*)}
\label{eleven:sec:memory-limited}

In problem domains where the rate of node generation by A* is high, 
the amount of memory available becomes a significant limitation, because A* can exhaust memory and terminate before finding a solution.
A*-based planners for domain-independent, classical planning such as Fast Downward \cite{helmert2006} generate between $10^4-10^5$ nodes per second on standard International Planning Competition benchmark domains. 
If a single state requires 100 bytes to represent, this means that A*-based planning can consume $10^6-10^7$ bytes per second.
Highly optimized solvers for specific domains such as the sliding-tile puzzle can generate over $10^6-10^7$ nodes per second \cite{burns2012implementing}, consuming memory even faster.
This problem is particularly pressing for parallel A* on a single machine.
Although the amount of RAM on a single machine has been steadily increasing, the number of cores on a single machine has also been rising, and the amount of memory \emph{per core} has remained fairly constant over the past decade (around 2GB/core).
If RAM is consumed at a rate of $10^7$ bytes per second, then A* will exhaust 1GB in approximately 100 seconds.
Thus, in domains with fast node generation rates, parallel A* can exhaust memory in a matter of minutes.

To overcome this limitation of A*, limited-memory, best-first search algorithms for finding optimal paths in implicit graphs have been extensively studied.
The best-known algorithm is Iterative Deepening A* (IDA*) \cite{korf:85a}. \index{iterative-deepening A* (IDA*)}
IDA* performs a series of depth-first searches, where each iteration is limited to an $f$-cost bound, which is increased on each iteration. 
Since each iteration of IDA* only performs a depth-first search, this requires memory which is only linear in the depth of the solution.
Although each iteration revisits all of the nodes visited on all of the previous iterations, many search spaces have the property that the runtime of iterative deepening is dominated by the search performed in the last few iterations, so the overhead of repeating the work done in past iterations is relatively small as a fraction of the total search effort \cite{korf:85a}.

However, if the search space is a graph, there may be many paths to each state, which results in significant amount of wasted search effort revisiting nodes through different paths.
To alleviate this problem with standard IDA*, a \emph{transposition table}, which is a cache of lower bounds on the solution cost achievable for previously visited states, can be added  so that search is pruned if a search reaches a previously visited state and it can be proven that the pruning does not result in loss of optimality \cite{reinefeldm94,akagikf2010}.
Other limited-memory A* variants include MA* \cite{ChakrabartiGAS89}, SMA* \cite{russell92}, and recursive best-first search \cite{korf:93}.

Below, we review parallel, limited-memory A* variants.

\subsection{Transposition Table-Driven Scheduling (TDS)}
\label{eleven:sec:tds}
Transposition-Table-Driven work scheduling (TDS)~\cite{Romein:99,Romein:2002} \index{transposition driven scheduling}
is a distributed-memory, parallel IDA* algorithm that uses a distributed transposition table.
Similarly to HDA*, TDS partitions the transposition table (TT) over processors and asynchronously distributes work using a Zobrist-based state hash function. 
In this way, TDS effectively allocates the large amount of distributed memory to the TT and uses the TT for efficiently detecting and pruning duplicate states that arrive at the processor. 
The distributed TT implementation uses the Zobrist hash function for mapping states to processors.
TDS initiates parallelism within each iteration and synchronizes between iterations. 
In a straightforward implementation of TDS, processors need to exchange messages that convey back-propagated lower bounds, but the efficient implementation of Romein et al.
eliminates such a back-propagation procedure, thus reducing communication overhead in exchange for giving up the use of more informed lower bounds (see \cite{Romein:99,Romein:2002} for details). 
Due to this modification, the Mattern's algorithm (Section \ref{eleven:sec:termination-detection}) is used for the termination detection of each IDA* iteration. 

Romein et al.~\cite{Romein:99,Romein:2002} showed that TDS exhibits a very low (sometimes negative) search overhead and yields significant (sometimes super-linear) speedups
in solving puzzles on a distributed-memory machine, compared to a sequential IDA* that runs on a single computational node with limited RAM capacity.
On the other hand, for domain-independent planning (International Planning Contest benchmark instances) , Kishimoto et al. showed that HDA* was consistently faster than TDS but sometimes terminates its execution due to memory exhaustion \cite{kishimotofb13}.  
Therefore, Kishimoto et al. proposed a simple, hybrid strategy combining HDA* and TDS. 
Their hybrid strategy first executes HDA* until either one of the HDA* processors exhausts its memory or the problem instance is solved. 
If HDA* fails to solve the problem instance, then TDS, which skips some wasteful iterations detected by
HDA* search, is executed. Thus, their hybrid strategy inherits advantages of both HDA* and TDS.

\subsection{Work Stealing for IDA*}

{\em Work stealing} \index{work stealing for parallel heuristic search}
is a standard approach for partitioning the search space, and is used particularly for parallelizing depth-first search in shared-memory environments.   
In work stealing, each processor maintains a local work queue.
When generating a new state, the processor places that state in its local queue. 
When the processor has no work in its queue, it ``steals'' work from the queue of a busy processor. 
Strategies for selecting a processor to steal the work from and determining the amount of work to steal 
are extensively studied (e.g. \cite{Rao:1987,Feldmann:1993,Frigo:98}). 

Nevertheless, work stealing suffers from performance degradation in domains where detecting duplicate states plays an important role. 
Romein et al. implemented work stealing for IDA* with transposition tables and compared it with TDS on a distributed-memory environment. 
They showed that TDS was 1.6 to 12.9 times faster than work stealing in puzzle solving domains \cite{Romein:2002}. 
On the other hand, TDS requires the use of a  reasonably a low-latency, high-bandwidth network for achieving efficient parallel performance.  
Therefore, Romein and Bal combined TDS with work stealing \cite{RomeinB01} in a grid environment where the communication latency is high between PC clusters and is low within each cluster. 
They use TDS to parallelize IDA* within each cluster for carrying out efficient duplication detection. When a cluster runs out of work, it steals work from another cluster, enabling much smaller communication overhead in the presence of the high-latency network.  

Variants of work stealing-based IDA* for Single Instruction, Multiple Data (SIMD) architecture machines have also been studied
\cite{powleyfk93,mahantid93}.
Since all processors in a SIMD machine must execute the same instruction, 
these approaches used a two-phase strategy which alternates between (1) a work (search) phase where all processors perform local, IDA* search, and (2) a load balancing phase, during which all processors exchange nodes.

To our knowledge, there is no published, empirical evaluation of work stealing for A* (as opposed to IDA*) in distributed-memory environments. This is a curious gap in the literature, given that work stealing is a standard approach for other parallel search models (e.g., branch-and-bound and backtracking for integer programming and  constraint programming).
This may be because work stealing strategies, particularly work stealing across machines in a cluster, tend to be more complex to implement than successful 
hash-based decentralized approaches such as HDA*. An investigation of work stealing approaches to A* therefore remains an avenue for future work.

\subsection{Parallel Window Search} 
\fromaij{}
Another approach to parallelizing IDA* is parallel-window search \cite{PowleyK91}, where
each processor searches from the same root node, \index{parallel window search}
but is assigned a different
bound -- that is, each processor is assigned a different, independent iteration of
IDA*. When a processor finishes an iteration, it is assigned the next
highest bound which has not yet been assigned to a processor.  
The
first solution found by parallel window IDA* is not necessarily
optimal. However, if, after finding a solution in the processor
assigned bound $b$, we wait until all processors with bound less than
$b$ finish, then the optimality of the best solution found is
assured.

\subsection{Parallel Retracting A* (PRA*)}
\label{eleven:sec:pra}

Parallel Retracting A* (PRA*) \cite{Evett:1995} \index{parallel retracting A* (PRA*)}
simultaneously addresses the problems of work distribution and duplicate state detection.
In PRA*, each processor maintains its own open and closed lists.
A hash function maps each state to exactly one processor which ``owns'' the state (as mentioned in Section \ref{eleven:sec:hash-based-decentralized}, the hash function used in PRA* was not specified in \cite{Evett:1995}).
When generating a state, PRA* distributes it to the corresponding owner.
If the hash keys are distributed uniformly across the processors, 
load balancing is achieved.
After receiving states, PRA* has the advantage that duplicate detection can be performed efficiently and locally at the destination processor.

While PRA* incorporated the idea of hash-based work distribution, PRA*
differs significantly from a parallel A* in that it is a parallel
version of Retracting A* (RA)* \cite{Evett:1995}, a limited-memory search algorithm closely related to MA*
\cite{ChakrabartiGAS89} and SMA* \cite{russell92}.  When a processor's
memory becomes full, Parallel Retracting A* {\em retracts} states from
the search frontier, and their $f$-values are stored in their parents,
which frees up memory.  Thus, unlike parallel A*, PRA* does not store
all expanded nodes in memory, and will not terminate due to running
out of memory in some process.  On the other hand, 
the implementation of this retraction mechanism in \cite{Evett:1995} incurs
a significant synchronization overhead:
when a processor $P$ generates a
new state $s$ and sends it to the destination processor $Q$, $P$
blocks and waits for $Q$ to confirm that $s$ has been successfully
received and stored (or whether the send operation failed due to
memory exhaustion at the destination processor).

%% file: astar-cloud.tex
\section{Parallel A* in Cloud Environments with Practically Unlimited Available Resources}
\label{eleven:sec:cloud}

Cloud computing resources such as Amazon EC2, which offer computational resources on demand, 
have become widely available in recent years.
In addition to cloud computing platforms, there is an increasing
availability of massive-scale, distributed grid computing resources
such as TeraGrid/XSEDE, as well as massively parallel,
high-performance computing (HPC) clusters.
These large-scale {\em utility computing resources} share two
characteristics that have significant implications for parallel search algorithms.
First, vast (practically unlimited) aggregate memory and CPU resources are available on demand. 
Secondly, resource usage incurs a direct monetary cost.

Previous work on parallel search algorithms has focused on
{\em makespan}: minimizing the runtime (wall-clock time) to find a solution, given fixed hardware resources; and 
{\em scalability}: as resource usage is increased, how are makespan and related metrics affected?
However, the availability of virtually unlimited resources at some cost introduces a new context for parallel search algorithm research 
where an explicit consideration of cost-performance  tradeoffs is necessary.
For scalable algorithms,
it is possible to reduce the makespan by allocating more
resources (up to some point). 
In practice, this incurs a high cost with diminishing marginal returns.
For parallel A* variants, under-allocating resources results in memory
exhaustion. On the other hand, over-allocation is costly and undesirable.
With the vast amounts of aggregate memory available in 
utility computing, the {\em cost}
(monetary funds) can be the new limiting factor, since one can
exhaust funds long before allocating all of the memory resources available from 
a large cloud service provider.

In utility computing services, 
there is some notion of an atomic unit of resource usage.
A \emph{hardware allocation unit} (HAU), is the minimal, discrete resource unit
that can be requested from a utility computing service.
Various HAU types can be available, each with different performance
characteristics and cost.
Commercial clouds such as EC2 tend to have
an immediate HAU allocation model with \emph{discrete} charges.
Usage of a HAU for any fraction of an hour is rounded up.
Grids and shared clusters tend to be batch-job based with a \emph{continuous}
cost model.
Jobs are submitted to a centralized
scheduler, with no
guarantees about when a job will be run.
The cost is a linear function of the amount of resources used.

\subsection{Iterative Allocation Strategy}

A scalable, {\em ravenous algorithm} is an algorithm that can run on an
arbitrary number of processors, and whose memory consumption
increases as it keeps running.
HDA* is an example of a scalable, ravenous algorithm.
The {\em iterative allocation} (IA) strategy \cite{FukunagaKB12}
repeatedly runs a ravenous algorithm $a$ until the problem is solved.
The key detail is deciding the number of HAUs to allocate in the next
iteration, if the previous iteration failed.
We seek a policy that tries to minimize the
total cost.


Two realistic assumptions which facilitate formal analysis are the following:
Firstly, all HAUs used by IA are identical hardware configurations.
Secondly, if a problem is solved on $i$ HAUs, then it will be solved
on $j > i$ HAUs (monotonicity).
Monotonicity is usually (implicitly) assumed in the previous work on parallel
search.
Let $T_v$ be the makespan (wall-clock) time needed to solve a problem on $v$
HAUs.
In a {continuous cost model},
the cost on $v$ HAUs is 
$T_v \times v$.
In a {discrete cost model},
the cost is $ \lceil T_v \rceil \times v$.
The {\em minimal width} $W^+$ is the minimum number of HAUs that can solve a problem 
with a given ravenous algorithm. 
Given a cost model (i.e., continuous or discrete),
$C^+$ is the associated \emph{min width cost}.
$C^*$ is the optimal cost to solve the problem, and  
the {\em minimal cost width} $W^*$ is the number of HAUs that results
in a minimal cost.
Since $W^*$ is usually not known a priori,  
the best we can hope for is to develop strategies that approximate the optimal values.
%
%
%
%
%

The {\em max iteration time} $E$ is the maximum actual (not rounded up) time
that an iteration can run 
before at least 1 HAU exhausts memory.

The \emph{min-width cost ratio} $R^+$ is defined as
$I(S)/C^+$, where $I(S)$ is the total cost of IA 
(using a particular allocation strategy $S$).
The \emph{min-cost ratio} $R^{*}$ is defined as
$I(S)/C^*$.
The total cost $I(S)$ of IA is accumulated over all iterations.
In a discrete cost model, times spent by individual HAUs
are rounded up.
The effect of the rounding up is alleviated by the fact that
HAUs will use any spare time left at the end of one iteration
to start the next iteration.


A particularly simple but useful strategy is the Geometric ($b^i$) Strategy, which was analyzed and evaluated by \cite{FukunagaKB12}.
The geometric strategy allocates $\lceil b^i \rceil$ HAUs
at iteration $i$, for some $b > 1$.
For example, the $2^i$ (doubling) strategy doubles the number
of HAUs allocated on each iteration.

Cloud platforms such as Amazon EC2 and Windows Azure typically have
discrete cost models, where the discrete billing unit is 1 hour.
This relatively long unit of time, combined
with the fast rate at which search algorithms consume RAM, leads to the
observation that
many (but not all) search applications will exhaust the RAM/core
in a HAU within a single billing time unit in modern cloud environments.
In other words, a single iteration of IA will complete (by either solving the
problem or running out of memory) within 1 billing time unit (i.e.,
$E \leq 1$).
%
%
This observation was experimentlly validated in \cite{FukunagaKB12}
for domain-independent planning benchmarks and 
 sequence alignment benchmarks.
In addition, HDA* has been observed to exhaust memory within 20 minutes on
every planning and 24-puzzle problem studied in \cite{kishimotofb13}.
With a sufficiently small $E$, 
{\em all iterations could be executed within a single billing time unit},
entirely eliminating the repeated allocation cost overhead.

In a discrete cost model with $E \leq 1$, the cost to
solve a problem on $v$ HAUs is proportional to $v$.
As a direct consequence, $W^+ = W^*$ and thus $R^+ = R^*$.
%
It can be shown that
in the {\em best case}, $R^*=R^+=1$,
in the {\em worst case},
$R_{wo}^* = R_{wo}^+ \leq \frac{b^2}{b-1}$,
and in the average case,
{ $R_{avg}^* = R_{avg}^+ \leq \frac{2b^2}{b^2-1}$}.
The worst case bound $b^2/(b-1)$ is minimized by the doubling
strategy ($b=2$).
As $b$ increases above 2, the upper bound for
$R^*_{avg}$ improves, 
but the worst case gets worse. 
Therefore, {\em the doubling strategy is the natural allocation policy to use in practice.}
%
For the $2^i$ strategy,
the average case ratio is bounded by $8/3 \approx 2.67$, 
and the worst case cost ratio does not exceed $4$.
With the $2^i$ strategy
in a discrete cost model when $E \leq 1$,
we \emph{never pay more than 4 times the optimal, but a priori unknown cost}.

%% file: astar-gpus.tex
\section{Parallel A* and IDA* on Graphics Processing Units}
\label{eleven:sec:gpu}

General-purpose computing using the thousands of cores available on Graphics Processing Units (GPUs) is currently a very active area of research. 
Zhou and Zeng propose a GPU-based A* algorithm using many (thousands) of parallel priority queues (OPEN lists)  \cite{zhou2015massively}.
A fundamental tradeoff successfully exploited by this approach is that by increasing the number of threads (parallel queues), they increase the effective parallelism. This results in duplicate node generations, but the duplicates  are efficiently detected and eliminated using hash-based duplicate detection.

The current bottleneck with executing A* entirely in the GPU is memory capacity -- the current, state-of-the-art GPU with the largest amount of RAM (Nvidia P100) has 16GB of global memory, which is an order of magnitude smaller than the amount of RAM on a current workstation. Since this GPU RAM is shared among thousands of cores, the amount of memory per core is several orders of magnitude smaller than the amount of RAM per core for the CPU, which limits the size of the search spaces that can be optimally searched.

As discussed in Section \ref{eleven:sec:memory-limited}, one approach to limit memory usage is iterative deepening. Horie and Fukunaga developed Block-Parallel IDA* (BPIDA*) \cite{HorieF17}, a parallel version of IDA* \cite{korf:85a} for the GPU. 
Although the single instruction, multi-thread architecture used in NVIDIA GPUs is somewhat similar to earlier SIMD architectures, Horie and Fukunaga found that simply porting earlier SIMD IDA* approaches \cite{powleyfk93,mahantid93} to the GPU results in extremely poor performance due to warp divergence and load balancing overheads. 
Instead of assigning a subtree of the search to a single thread as SIMD IDA* does, BPIDA* assigns a subtree to a GPU block (a group of threads which execute on the same streaming multiprocessor and share memory), and each block has a shared, parallel open list. This was shown to significantly improve parallel efficiency on the 15-puzzle.
Their implementation of BPIDA* only uses the shared memory, and completely avoids using the GPU global memory (RAM on the GPU which is shared by all streaming multiprocessors). This was possible because 15-puzzle states can be represented compatly enough that the search stacks fit entirely in shared memory; in addition, they used Mahnattan distance as the heuristic function, which requires no memory.
Thus, BPIDA* achieves good parallel efficiency but the search is not efficient compared to a state-of-the-art IDA* implementation which uses
a more powerful but memory-intensive heuristic function (e.g., pattern databases \cite{korff02}). 
Using such memory-intensive heuristics (as well as other memory-intensive methods such as a transposition tables \cite{reinefeldm94}) on the GPU will require using the global memory and is a direction for future work.

Heterogeneous approaches which use both the GPU as well as CPU is an open area for future work.
One instance of such a hybrid GPU/CPU based approach is for best-fist search with a blind heuristic by Sulweski et al \cite{SulewskiEK11}. Their algorithm uses a GPU to accelerate precondition checks and successor generation, but uses the CPU for duplicate detection.

Finally, a different application of many-core GPU architectures is for multi-agent search, where each core executes an independent A* search for each agent in the simulation environment  \cite{Bleiweiss08}.

%% file: other-approaches.tex
\section{Other Approaches}
\label{eleven:sec:misc-approaches}

One alternative to partitioning the search space among processors is to parallelize the computation done during the
processing of a single search node (cf., \cite{CampbellHH02, CazenaveJ08}).  
%
The Operator Distribution Method for parallel Planning (ODMP) \cite{VrakasRV01}
parallelizes the computation at each node.
In ODMP, there is a single {\em
  controlling thread}, and several {\em planning threads}.  The controlling
thread is responsible for initializing and maintaining the current
search state. At each step of the controlling-thread main loop, 
it generates the applicable operators, inserts them in
an {\em operator pool}, and activates the planning threads. Each
planning thread independently takes an operator from this shared
operator pool, computes the grounded actions, generates the
resulting states, evaluates the states with the heuristic function,
and stores the new state and its heuristic value in a {\em global agenda} 
data structure. 
After the operator pool is empty,  
the controlling thread extracts the best new
state from the global agenda, assigning it to the new, current
state. 



The best parallelization strategy for a search algorithm depends on
the properties of the search space, as well as the parallel architecture on
which the search algorithm is executed. The EUREKA system
\cite{CookV98} used machine learning to automatically configure
parallel IDA* for various problems (including nonlinear planning) and machine architectures.

\todoadi{-- need more some more details, and maybe needs to be moved to (sub)section with other related approaches -- I think Adi read this paper for the AIJ paper?}
Niewiadomski et al. \cite{Niewiadomski06} propose PFA*-DDD, a parallel version of 
Frontier A* with Delayed Duplicate Detection. PFA*-DDD partitions the open sets into groups (interval lists) and assigns them to processors. 
PFA*-DDD returns the cost of a path from start to target,
not an actual path. 
While divide-and-conquer (DC) can be used to reconstruct a path (as in sequential frontier search), parallel DC poses non-trivial design issues that need to be addressed in future work. 



%% file: parallel-astar.bbl
\begin{thebibliography}{10}
\providecommand{\url}[1]{{#1}}
\providecommand{\urlprefix}{URL }
\expandafter\ifx\csname urlstyle\endcsname\relax
  \providecommand{\doi}[1]{DOI~\discretionary{}{}{}#1}\else
  \providecommand{\doi}{DOI~\discretionary{}{}{}\begingroup
  \urlstyle{rm}\Url}\fi

\bibitem{akagikf2010}
Akagi, Y., Kishimoto, A., Fukunaga, A.: On transposition tables for
  single-agent search and planning: Summary of results.
\newblock In: Proceedings of the 3rd Symposium on Combinatorial Search (SOCS),
  pp. 1--8 (2010)

\bibitem{backstrom1995complexity}
B{\"a}ckstr{\"o}m, C., Nebel, B.: Complexity results for {SAS}+ planning.
\newblock Computational Intelligence \textbf{11}(4), 625--655 (1995)

\bibitem{Bleiweiss08}
Bleiweiss, A.: {GPU} accelerated pathfinding.
\newblock In: Proceedings of the {EUROGRAPHICS/ACM} {SIGGRAPH} Conference on
  Graphics Hardware 2008, Sarajevo, Bosnia and Herzegovina, 2008, pp. 65--74
  (2008).
\newblock \doi{10.2312/EGGH/EGGH08/065-074}.
\newblock \urlprefix\url{http://dx.doi.org/10.2312/EGGH/EGGH08/065-074}

\bibitem{burnslrz10}
Burns, E., Lemons, S., Ruml, W., Zhou, R.: Best-first heuristic search for
  multicore machines.
\newblock Journal of Artificial Intelligence Research (JAIR) \textbf{39},
  689--743 (2010)

\bibitem{Burns:09}
Burns, E., Lemons, S., Zhou, R., Ruml, W.: Best-first heuristic search for
  multi-core machines.
\newblock In: Proceedings of the Twenty-First International Joint Conference on
  Artificial Intelligence IJCAI-09 (2009)

\bibitem{burns2012implementing}
Burns, E.A., Hatem, M., Leighton, M.J., Ruml, W.: Implementing fast heuristic
  search code.
\newblock pp. 25--32 (2012)

\bibitem{CampbellHH02}
Campbell, M., Hoane, J., Hsu, F.: Deep blue.
\newblock Artificial Intelligence \textbf{134}(1-2), 57--83 (2002)

\bibitem{CazenaveJ08}
Cazenave, T., Jouandeau, N.: On the parallelization of {UCT}.
\newblock In: H.~van den Herik~et al. (ed.) Proceedings of Computers and Games
  CG-08, \emph{LNCS}, vol. 5131, pp. 72--80. Springer (2008)

\bibitem{ChakrabartiGAS89}
Chakrabarti, P., Ghose, S., Acharya, A., de~Sarkar, S.: Heuristic search in
  restricted memory.
\newblock Artificial Intelligence \textbf{41}(2), 197--221 (1989)

\bibitem{CookV98}
Cook, D., Varnell, R.: Adaptive parallel iterative deepening search.
\newblock Journal of Artificial Intelligence Research \textbf{9}, 139--166
  (1998)

\bibitem{cormen01}
Cormen, T.H., Leiserson, C.E., Rivest, R.L., Stein, C.: Introduction to
  Algorithms, Second Edition.
\newblock The MIT Press (2001).
\newblock
  \urlprefix\url{http://www.amazon.ca/exec/obidos/redirect?tag=citeulike09-20{\&}path=ASIN/0262531968}

\bibitem{DuttM94}
Dutt, S., Mahapatra, N.: Scalable load balancing strategies for parallel {A*}
  algorithms.
\newblock Journal of parallel and distributed computing \textbf{22}, 488--505
  (1994)

\bibitem{Edelkamp:2010:HST:1875144}
Edelkamp, S., Schroedl, S.: Heuristic Search: Theory and Applications.
\newblock Morgan Kaufmann Publishers Inc., San Francisco, CA, USA (2010)

\bibitem{Evett:1995}
Evett, M., Hendler, J., Mahanti, A., Nau, D.: $\mbox{PRA}^*$: Massively
  parallel heuristic search.
\newblock Journal of Parallel and Distributed Computing \textbf{25}(2),
  133--143 (1995)

\bibitem{Feldmann:1993}
Feldmann, R.: Spielbaumsuche auf massiv parallelen systemen.
\newblock Ph.D. thesis, University of Paderborn (1993).
\newblock English translation titled {\it {Game tree search on massively
  parallel systems}} is available.

\bibitem{Felner03kbfs:k-best-first}
Felner, A., Kraus, S., Korf, R.E.: Kbfs: K-best-first search.
\newblock Annals of Mathematics and Artificial Intelligence \textbf{39}, 19--39
  (2003)

\bibitem{Frigo:98}
Frigo, M., Leiserson, C.E., Randall, K.H.: The implementation of the cilk-5
  multithreaded language.
\newblock In: ACM SIGPLAN Conferences on Programming Language Design and
  Implementation (PLDI'98), pp. 212--223 (1998)

\bibitem{FukunagaKB12}
Fukunaga, A., Kishimoto, A., Botea, A.: Iterative resource allocation for
  memory intensive parallel search algorithms on clouds, grids, and shared
  clusters.
\newblock In: Proceedings of the National Conference on Artificial %
  Intelligence (AAAI) (2012).
\newblock
  \urlprefix\url{http://www.aaai.org/ocs/index.php/AAAI/AAAI12/paper/view/5054}

\bibitem{furcy:05}
Furcy, D., Koenig, S.: Limited discrepancy beam search.
\newblock In: Proceedings of the International Joint Conference on Artificial
  Intelligence, pp. 125--131 (2005)

\bibitem{GaravelMS01}
Garavel, H., Mateescu, R., Smarandache, I.M.: Parallel state space construction
  for model-checking.
\newblock In: Proceedings of the 8th International SPIN Workshop, pp. 217--234
  (2001)

\bibitem{GomesSCK00}
Gomes, C., Selman, B., Crato, N., Kautz, H.: Heavy-tailed phenomena in
  satisfiability and constraint satisfaction problems.
\newblock Journal of Automated Reasoning \textbf{24}(1-2), 67--100 (2000)

\bibitem{HamadiJS09}
Hamadi, Y., Jabbour, S., Sais, L.: {ManySAT}: a parallel {SAT} solver.
\newblock Journal on Satisfiability, Boolean Modeling and Computation
  \textbf{6}, 245--262 (2009)

\bibitem{hart68formal}
Hart, P., Nilsson, N., Raphael, B.: A formal basis for the heuristic
  determination of minimum cost paths.
\newblock IEEE Transactions on System Sciences and Cybernetics
  \textbf{SSC-4}(2), 100--107 (1968)

\bibitem{helmert2006}
Helmert, M.: The {Fast Downward} planning system.
\newblock Journal of Artificial Intelligence Research \textbf{26}, 191--246
  (2006).
\newblock \doi{10.1613/jair.1705}

\bibitem{Helmert:07}
Helmert, M., Haslum, P., Hoffmann, J.: Flexible abstraction heuristics for
  optimal sequential planning.
\newblock In: Proceedings of the Seventeenth International Conference on
  Automated Planning and Scheduling ICAPS-07, pp. 176--183 (2007)

\bibitem{DBLP:conf/cav/HeymanGGS00}
Heyman, T., Geist, D., Grumberg, O., Schuster, A.: Achieving scalability in
  parallel reachability analysis of very large circuits.
\newblock In: Proceedings 12th International Conference on Computer Aided
  Verification, pp. 20--35 (2000)

\bibitem{HolzmannB07}
Holzmann, G.J., Bo\^{s}na\^{c}ki, D.: The design of a multicore extension of
  the {SPIN} model checker.
\newblock IEEE Transactions on Software Engineering \textbf{33}(10), 659--674
  (2007)

\bibitem{HorieF17}
Horie, S., Fukunaga, A.S.: Block-parallel ida* for gpus.
\newblock In: Proceedings of the Tenth International Symposium on Combinatorial
  Search, Edited by Alex Fukunaga and Akihiro Kishimoto, 16-17 June 2017,
  Pittsburgh, Pennsylvania, {USA.}, pp. 134--138 (2017).
\newblock
  \urlprefix\url{https://aaai.org/ocs/index.php/SOCS/SOCS17/paper/view/15801}

\bibitem{HubermanLH97}
Huberman, B., Lukose, R., Hogg, T.: An economics approach to hard computational
  problems.
\newblock Science \textbf{275}(5296), 51--54 (1997)

\bibitem{iranis86}
Irani, K., Shih, Y.: Parallel {A*} and {AO*} algorithms: An optimality
  criterion and performance evaluation.
\newblock In: International Conference on Parallel Processing, pp. 274--277
  (1986)

\bibitem{jinnai2016structured}
Jinnai, Y., Fukunaga, A.: Abstract {Z}obrist hashing: An efficient work
  distribution method for parallel best-first search.
\newblock In: Proceedings of the National Conference on Artificial %
  Intelligence (AAAI), pp. 717--723 (2016)

\bibitem{jinnai2016automated}
Jinnai, Y., Fukunaga, A.: Automated creation of efficient work distribution
  functions for parallel best-first search.
\newblock In: {Proc. ICAPS} (2016)

\bibitem{jinnai2017work}
Jinnai, Y., Fukunaga, A.: On work distribution functions for parallel
  best-first search.
\newblock Journal of Artificial Intelligence Research  (2017).
\newblock (to appear)

\bibitem{karpz88}
Karp, R., Zhang, Y.: A randomized parallel branch-and-bound procedure.
\newblock In: Proceedings of the 20th {ACM} Symposium on Theory of Computing
  (STOC), pp. 290--300 (1988)

\bibitem{karpz93}
Karp, R., Zhang, Y.: Randomized parallel algorithms for backtrack search and
  branch-and-bound computation.
\newblock Journal of the Association for Computing Machinery \textbf{40}(3),
  765--789 (1993)

\bibitem{kishimotofb13}
Kishimoto, A., Fukunaga, A., Botea, A.: Evaluation of a simple, scalable,
  parallel best-first search strategy.
\newblock Artificial Intelligence \textbf{195}, 222--248 (2013).
\newblock \doi{10.1016/j.artint.2012.10.007}.
\newblock
  \urlprefix\url{http://linkinghub.elsevier.com/retrieve/pii/S0004370212001294}

\bibitem{kishimotofb09}
Kishimoto, A., Fukunaga, A.S., Botea, A.: Scalable, parallel best-first search
  for optimal sequential planning.
\newblock In: {Proc. ICAPS}, pp. 201--208 (2009).
\newblock
  \urlprefix\url{http://aaai.org/ocs/index.php/ICAPS/ICAPS09/paper/view/705}

\bibitem{Knight93}
Knight, K.: Are many reactive agents better than a few deliberative ones?
\newblock In: Proceedings of the 13th International Joint Conference on
  Artificial Intelligence, pp. 432--437 (1993)

\bibitem{knuth73}
Knuth, D.E.: "Sorting and Searching", The Art of Computer Programming, vol.~3.
\newblock Addison-Wesley (1973)

\bibitem{KobayashiKW2011}
Kobayashi, Y., Kishimoto, A., Watanabe, O.: Evaluations of {Hash} {Distributed}
  {A*} in optimal sequence alignment.
\newblock In: Proceedings of the 22nd International Joint Conference on
  Artificial Intelligence, pp. 584--590 (2011)

\bibitem{korf:85a}
Korf, R.: Depth-first iterative deepening: An optimal admissible tree search.
\newblock Artificial Intelligence \textbf{97}, 97--109 (1985)

\bibitem{korf:93}
Korf, R.: {Linear-Space Best-First Search}.
\newblock Artificial Intelligence \textbf{62}(1), 41--78 (1993)

\bibitem{korff02}
Korf, R.E., Felner, A.: Disjoint pattern database heuristics.
\newblock Artificial Intelligence \textbf{134}(1-2), 9--22 (2002)

\bibitem{Korf:2000}
Korf, R.E., Zhang, W.: Divide-and-conquer frontier search applied to optimal
  sequence alignment.
\newblock In: Proceedings of the 17th National Conference on Artificial
  Intelligence AAAI-00, pp. 910--916 (2000)

\bibitem{Kumar05loadbalancing}
Kumar, R., Mercer, E.G.: Load balancing parallel explicit state model checking.
\newblock Electronic Notes in Theoretical Computer Science \textbf{128} (2005)

\bibitem{Kumar:1988}
Kumar, V., Ramesh, K., Rao, V.N.: Parallel best-first search of state-space
  graphs: A summary of results.
\newblock In: Proceedings of the 7th National Conference on Artificial
  Intelligence AAAI-88, pp. 122--127 (1988)

\bibitem{kumar1988parallel}
Kumar, V., Ramesh, K., Rao, V.N.: Parallel best-first search of state-space
  graphs: A summary of results.
\newblock In: Proceedings of the National Conference on Artificial %
  Intelligence (AAAI), vol.~88, pp. 122--127 (1988)

\bibitem{lerda-99}
Lerda, F., Sisto, R.: Distributed-memory model checking with {SPIN}.
\newblock In: Theoretical and Practical Aspects of SPIN Model Checking, 5th and
  6th International SPIN Workshops, \emph{Lecture Notes in Computer Science},
  vol. 1680, pp. 22--39 (1999)

\bibitem{mahantid93}
Mahanti, A., Daniels, C.: A {SIMD} approach to parallel heuristic search.
\newblock Artificial Intelligence \textbf{60}, 243--282 (1993)

\bibitem{MahapatraD97}
Mahapatra, N., Dutt, S.: Scalable global and local hashing strategies for
  duplicate pruning in parallel {A*} graph search.
\newblock IEEE Transactions on Parallel and Distributed Systems \textbf{8}(7),
  738--756 (1997)

\bibitem{Mattern:87}
Mattern, F.: Algorithms for distributed termination detection.
\newblock Distributed Computing \textbf{2}(3), 161--175 (1987)

\bibitem{Melatti:2009}
Melatti, I., Palmer, R., Sawaya, G., Yang, Y., Kirby, R.M., Gopalakrishnan, G.:
  Parallel and distributed model checking in {Eddy}.
\newblock International Journal on Software Tools for Technology Transfer
  \textbf{11}(1), 13--25 (2009)

\bibitem{Niewiadomski06}
Niewiadomski, R., Amaral, J.N., Holte, R.C.: Sequential and parallel algorithms
  for frontier {A*} with delayed duplicate detection.
\newblock In: Proceedings of the 21st National Conference on Artificial
  Intelligence (AAAI), pp. 1039--1044 (2006)

\bibitem{Pearl84}
Pearl, J.: Heuristics - Intelligent Search Strategies for Computer Problem
  Solving.
\newblock Addison--Wesley (1984)

\bibitem{PhillipsLK14}
Phillips, M., Likhachev, M., Koenig, S.: {PA*SE}: Parallel {A*} for slow
  expansions.
\newblock In: {Proc. ICAPS} (2014).
\newblock
  \urlprefix\url{http://www.aaai.org/ocs/index.php/ICAPS/ICAPS14/paper/view/7952}

\bibitem{powleyfk93}
Powley, C., Ferguson, C., Korf, R.: Depth-first heuristic search on a {SIMD}
  machine.
\newblock Artificial Intelligence \textbf{60}, 199--242 (1993)

\bibitem{PowleyK91}
Powley, C., Korf, R.: Single-agent parallel window search.
\newblock {IEEE} Transactions on Pattern Analysis and Machine Intelligence
  \textbf{13}(5), 466--477 (1991)

\bibitem{Rao:1987}
Rao, V.N., Kumar, V.: Parallel depth-first search on multiprocessors part {I}:
  Implementation.
\newblock International Journal of Parallel Programming \textbf{16}(6),
  479--499 (1987)

\bibitem{reinefeldm94}
Reinefeld, A., Marsland, T.: Enhanced iterative-deepening search.
\newblock IEEE Transactions on Pattern Analysis and Machine Intelligence
  \textbf{16}(7), 701--710 (1994)

\bibitem{RomeinB01}
Romein, J.W., Bal, H.E.: Wide-area transposition-driven scheduling.
\newblock In: Proceedings of the 10th IEEE International Symposium on High
  Performance Distributed Computing, pp. 347--355 (2001)

\bibitem{Romein:2002}
Romein, J.W., Bal, H.E., Schaeffer, J., Plaat, A.: A performance analysis of
  transposition-table-driven work scheduling in distributed search.
\newblock IEEE Transactions on Parallel and Distributed Systems \textbf{13}(5),
  447--459 (2002)

\bibitem{romein1999transposition}
Romein, J.W., Plaat, A., Bal, H.E., Schaeffer, J.: Transposition table driven
  work scheduling in distributed search.
\newblock In: Proceedings of the National Conference on Artificial %
  Intelligence (AAAI), pp. 725--731 (1999)

\bibitem{Romein:99}
Romein, J.W., Plaat, A., Bal, H.E., Schaeffer, J.: Transposition table driven
  work scheduling in distributed search.
\newblock In: Proceedings of the National Conference on Artificial Intelligence
  AAAI-99, pp. 725--731 (1999)

\bibitem{russell92}
Russell, S.: Efficient memory-bounded search methods.
\newblock In: Proc. ECAI (1992)

\bibitem{SternD97}
Stern, U., Dill, D.L.: Parallelizing the {Murphi} verifier.
\newblock In: Proceedings of the 9th International Conference on Computed Aided
  Verification, pp. 256--278 (1997)

\bibitem{SternD01}
Stern, U., Dill, D.L.: Parallelizing the {Murphi} verifier.
\newblock Formal Methods in System Design \textbf{18}(2), 117--129 (2001)

\bibitem{SulewskiEK11}
Sulewski, D., Edelkamp, S., Kissmann, P.: Exploiting the computational power of
  the graphics card: Optimal state space planning on the {GPU}.
\newblock In: Proceedings of the 21st International Conference on Automated
  Planning and Scheduling, {ICAPS} 2011, Freiburg, Germany June 11-16, 2011
  (2011).
\newblock
  \urlprefix\url{http://aaai.org/ocs/index.php/ICAPS/ICAPS11/paper/view/2699}

\bibitem{ValenzanoNMSS12}
Valenzano, R., Nakhost, H., M\"uller, M., Schaeffer, J., Sturtevant, N.:
  Arvandherd: Parallel planning with a portfolio.
\newblock In: Proceedings of the 20th European Conference on Artificial
  Intelligence, pp. 786--791 (2012)

\bibitem{ValenzanoSSBK10}
Valenzano, R., Sturtevant, N., Schaeffer, J., Buro, K., Kishimoto, A.:
  Simultaneously searching with multiple settings: An alternative to parameter
  tuning for suboptimal single-agent search algorithms.
\newblock In: Proceedings of the 20th International Conference on Automated
  Planning and Scheduling, pp. 177--184 (2010)

\bibitem{VBH:socs2010}
Vidal, V., Bordeaux, L., Hamadi, Y.: Adaptive k-parallel best-first search: A
  simple but efficient algorithm for multi-core domain-independent planning.
\newblock In: Proceedings of the 3rd Symposium on Combinatorial Search
  (SOCS'10) (2010)

\bibitem{VrakasRV01}
Vrakas, D., Refanidis, I., Vlahavas, I.: Parallel planning via the distribution
  of operators.
\newblock Journal of Experimental and Theoretical Artificial Intelligence
  \textbf{13}(3), 211--226 (2001)

\bibitem{Zhou:2006}
Zhou, R., Hansen, E.: Domain-independent structured duplicate detection.
\newblock In: Proceedings of the 21st National Conference on Artificial
  Intelligence AAAI-06, pp. 683--688 (2006)

\bibitem{zhou2004structured}
Zhou, R., Hansen, E.A.: Structured duplicate detection in external-memory graph
  search.
\newblock In: Proceedings of the National Conference on Artificial %
  Intelligence (AAAI), pp. 683--689 (2004)

\bibitem{zhou2006breadth}
Zhou, R., Hansen, E.A.: Breadth-first heuristic search.
\newblock Artificial Intelligence \textbf{170}(4), 385--408 (2006)

\bibitem{zhou2006}
Zhou, R., Hansen, E.A.: Domain-independent structured duplicate detection.
\newblock In: Proceedings of the National Conference on Artificial %
  Intelligence (AAAI), pp. 1082--1087 (2006)

\bibitem{zhou2007parallel}
Zhou, R., Hansen, E.A.: Parallel structured duplicate detection.
\newblock In: Proceedings of the National Conference on Artificial %
  Intelligence (AAAI), pp. 1217--1223 (2007)

\bibitem{zhou2015massively}
Zhou, Y., Zeng, J.: Massively parallel {A}* search on a {GPU}.
\newblock In: Proceedings of the National Conference on Artificial %
  Intelligence (AAAI), pp. 1248--1255 (2015).
\newblock
  \urlprefix\url{http://www.aaai.org/ocs/index.php/AAAI/AAAI15/paper/view/9620}

\bibitem{Zobrist1970}
Zobrist, A.L.: A new hashing method with application for game playing.
\newblock {reprinted in} {I}nternational Computer Chess Association Journal
  ({ICCA}) \textbf{13}(2), 69--73 (1970)

\end{thebibliography}
